\documentclass[jair,11pt,theapa,oneside]{article}
\usepackage[hidelinks]{hyperref}

\hypersetup{
pdftitle={A Survey of Zero-shot Generalisation in Deep Reinforcement Learning},
pdfsubject={cs.AI, cs.LG},
pdfauthor={Robert Kirk, Amy Zhang,  Edward Grefenstette, Tim Rockt\"{a}schel},
pdfkeywords={Generalisation, Reinforcement Learning, Survey, Review},
colorlinks=true,
linkcolor=blue,
urlcolor=blue,
}

\usepackage{jair, theapa, rawfonts}

\usepackage[dvipsnames]{xcolor}
\usepackage[utf8]{inputenc} 
\usepackage[T1]{fontenc}    
\usepackage{url}            
\usepackage{amsfonts}       
\usepackage{nicefrac}       
\usepackage{microtype}      
\usepackage{nameref}
\usepackage{color, colortbl}
\usepackage{amsmath}
\usepackage{amsthm}
\usepackage{booktabs}
\usepackage[font=small]{caption}
\usepackage{svg}
\usepackage{lscape}
\usepackage{graphicx}
\usepackage{array}
\usepackage{tabularx}
\usepackage{soul}
\usepackage{doi}
\usepackage{longtable}
\usepackage{moresize}
\usepackage[capitalise]{cleveref}
\usepackage{xcolor, soul}

\graphicspath{{images/}}

\jairheading{76}{2023}{201-264}{08/2022}{01/2023}
\ShortHeadings{A Survey of Zero-shot Generalisation in Deep Reinforcement Learning}
{Kirk, Zhang, Grefenstette, \& Rockt\"{a}schel}
\firstpageno{201}

\title{A Survey of Zero-shot Generalisation in Deep \\Reinforcement Learning}

\author{\name Robert Kirk \email robert.kirk.20@ucl.ac.uk \\
	\addr University College London, Gower St, London\\WC1E 6BT, United Kingdom
	\AND
    \name Amy Zhang \email amyzhang@fb.com \\
	\addr University of California, Berkeley, Berkeley\\CA, United States \\Meta AI Research, 
	\AND
	\name Edward Grefenstette \email e.grefenstette@ucl.ac.uk \\
	\addr University College London, Gower St, London\\WC1E 6BT, United Kingdom
    \AND
	\name Tim Rockt\"{a}schel \email tim.rocktaschel@ucl.ac.uk \\
	\addr University College London, Gower St, London\\WC1E 6BT, United Kingdom \\
}

\date{}

\newtheorem{definition}{Definition}

\begin{document}
\maketitle

\renewcommand{\thefootnote}{\arabic{footnote}}
\newcommand{\R}{\mathbb{R}}

\definecolor{Gray}{gray}{0.9}
\definecolor{CBBlue}{RGB}{0,114,178}
\definecolor{CBGreen}{RGB}{0,158,115}
\definecolor{CBOrange}{RGB}{213,94,0}
\definecolor{CBPink}{RGB}{204,121,167}

\begin{abstract} 
The study of zero-shot generalisation (ZSG) in deep Reinforcement Learning (RL) aims to produce RL algorithms whose policies generalise well to novel unseen situations at deployment time, avoiding overfitting to their training environments. Tackling this is vital if we are to deploy reinforcement learning algorithms in real world scenarios, where the environment will be diverse, dynamic and unpredictable. This survey is an overview of this nascent field. We rely on a unifying formalism and terminology for discussing different ZSG problems, building upon previous works. We go on to categorise existing benchmarks for ZSG, as well as current methods for tackling these problems. Finally, we provide a critical discussion of the current state of the field, including recommendations for future work. Among other conclusions, we argue that taking a purely procedural content generation approach to benchmark design is not conducive to progress in ZSG, we suggest fast online adaptation and tackling RL-specific problems as some areas for future work on methods for ZSG, and we recommend building benchmarks in underexplored problem settings such as offline RL ZSG and reward-function variation.
\end{abstract}
\section{Introduction}\label{section:intro}

Reinforcement Learning (RL) has the potential to be used in a wide range of applications from autonomous vehicles \shortcite{filosCanAutonomousVehicles2020} and algorithm control \shortcite{biedenkappDynamicAlgorithmConfiguration2020a} to robotics \shortcite{openaiSolvingRubikCube2019}, but to fulfil this potential we need RL algorithms that can be used in the real world. Reality is dynamic, open-ended and always changing, and RL algorithms will need to be robust to variations in their environments, and have the capability to transfer and adapt to unseen (but similar) environments during their deployment.

\begin{figure}[t]
\centering
\includegraphics[width = \textwidth]{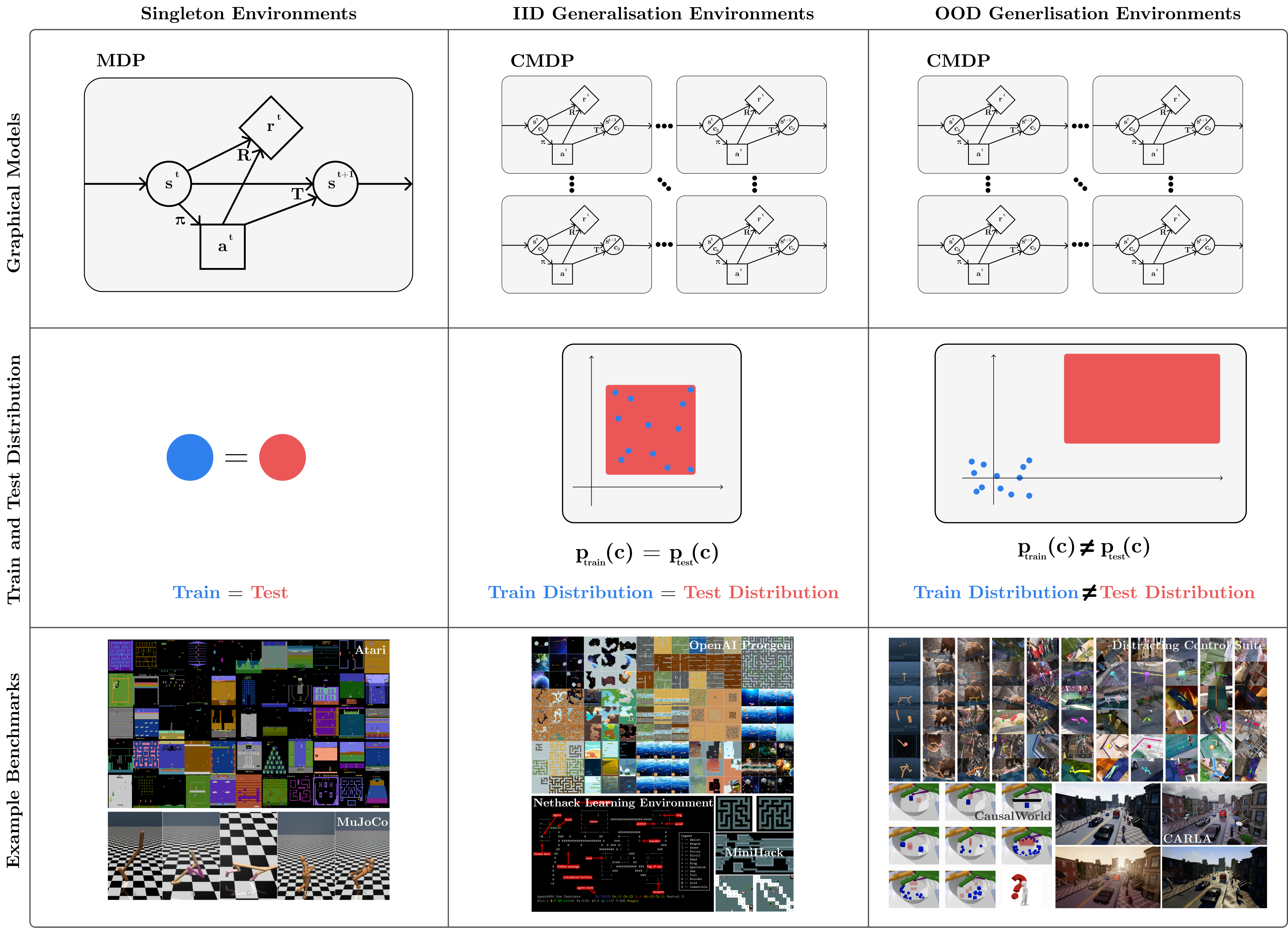}
\caption{\textbf{Zero-shot Generalisation in Reinforcement Learning}. A visualisation of three types of environment (columns) with respect to their graphical model, training and testing distribution and example benchmarks (rows). Classical RL has focused on environments where training and testing are identical (singleton environments, first column). We focus on an underexplored setting, inspired by likely real-world scenarios, where training and testing environments will be different, with environment instances either from the same distribution (Independent and Identically Distributed (IID) ZSG Environments, second column) or from different distributions (OOD ZSG Environments, third column). The split between the second and third columns is just one example of the way in which ZSG is a class of problems rather than an individual problem. The top row visualises the differences in graphical models between singleton environments and environments where ZSG is required. For more information on the CMDP formalism see \protect{\cref{subsection:cmdp}}.}\label{fig:leadvis}
\end{figure}

However, much current RL research works on benchmarks such as Atari \shortcite{bellemareArcadeLearningEnvironment2013} and MuJoCo \shortcite{e.todorovMuJoCoPhysicsEngine2012,brockmanOpenAIGym2016}, which do not have the attributes described above: they evaluate the policy on exactly the same environment it was trained on, which often does not match with real-world scenarios (\cref{fig:leadvis} left column). This is in stark contrast to the standard assumptions of supervised learning where the training and testing sets are disjoint and is likely to lead to strong evaluation overfitting \shortcite{whitesonProtectingEvaluationOverfitting2011}. This has resulted in policies that perform badly on even slightly adjusted environment instances (specific levels or tasks within an environment) and often fail on unseen random seeds used for initialisation \shortcite{zhangDissectionOverfittingGeneralization2018,zhangStudyOverfittingDeep2018,farebrotherGeneralizationRegularizationDQN2020,gamrianTransferLearningRelated2019}.

In this work we survey the recent literature studying zero-shot generalisation in deep RL, a field focused on producing algorithms with the robustness, transfer and adaptation properties required to perform well in the real world. We offer a unifying framework that builds on previous work \shortcite{ghoshWhyGeneralizationRL2021,hallakContextualMarkovDecision2015,harrisonADAPTZeroShotAdaptive2017,higginsDARLAImprovingZeroShot2018,songObservationalOverfittingReinforcement2019,perezGeneralizedHiddenParameter2020} that formalises the problem of ZSG in RL as a \emph{class} of problems, rather than a single problem. 
While prior work \shortcite{ghoshWhyGeneralizationRL2021,hallakContextualMarkovDecision2015,perezGeneralizedHiddenParameter2020} uses the contextual MDP framework or related frameworks to describe how an agent can encounter new, unseen states at test time to generalise to, we further extend and break down the types of generalisation that can be possible, e.g.~combinatorial, interpolation vs.~extrapolation, single-factor vs.~multi-factor (\cref{fig:evalprotcontrol}). We formalise more fully the ZSG problem, formally specifying the policy class, making clear the choice of whether the context is observed or not, and including cases where the context distribution is controllable during training \cref{def:zsptcc}
This breakdown enables us to clearly and crisply compare previous works, as well as understand how to choose future research directions. For example, improving ``generalisation'' without any additional assumptions is inherently underspecified; it is unlikely that we can generically improve generalisation, given this class of problems is so broad that some analogy of the No Free Lunch theorem \shortcite{wolpertNoFreeLunch1997} applies: improving generalisation in some settings could harm generalisation in others.
Two broad categories of ZSG problem are shown in \cref{fig:leadvis} in the centre and right columns.

Using this formalism, we survey and examine the range of benchmarks available for ZSG in RL, and go on to discuss methods aimed at tackling different ZSG problems. Finally, we propose several settings within ZSG which are underexplored but still vital for various real-world applications of RL, as well as many avenues for future work on methods that can solve different generalisation problems. Throughout, we critically review the state of the field and provide recommendations for ensuring future research is robust and useful. We aim to make the field more legible to researchers and practitioners both in and out of the field and make discussing new research directions easier by providing a common reference and framework. This new clarity can improve the field, and enable robust progress towards more general RL methods.

\paragraph{Scope.} Generalisation in RL captures a large amount of research, so to make this survey feasible we limit the scope of our review in several ways. First, we focus on the specific problem setting of zero-shot generalisation (ZSG), where the policy is evaluated zero-shot on a collection of environment instances different to those it was trained on. Crucially, this setting disallows any additional training in or data from the test environment instances, meaning methods such as domain adaptation and many meta-RL approaches are not applicable. This is different from classical RL, which has historically focused on online learning in a single MDP, where generalisation refers to the notion of generalising to novel states in the same MDP. While this setting can be useful to study, we instead focus on situations which require policies that can be deployed in situations they haven't been trained in, and generalise well zero-shot to those situations. This setting is especially relevant for current deep RL algorithms, which often aren't sample-efficient or safe enough to be deployed online without significant offline or in-simulation training first. This motivates our focus on ZSG\@. We discuss and motivate this setting and its restrictions more in \cref{paragraph:0shottransfer}.

Second, we only cover single-agent RL in this work. There are generalisation problems within multi-agent reinforcement learning (MARL), such as being general enough to defeat multiple different opponent strategies \shortcite{openaiDotaLargeScale2019,vinyalsGrandmasterLevelStarCraft2019} and generalising to new team-mates in cooperative games \shortcite{huOffBeliefLearning2021,huOtherPlayZeroShotCoordination2021}, but we do not cover any work in this area here. While mathematically these problems could be modelled equivalently (if co-players are modelled as parts of the dynamics function, rather than as agents as in Games-based formulations \shortcite{shapleyStochasticGames1953}), approaches to these problems tend to be quite different, explicitly utilising the fact that variation and generalisation challenges come from other co-players rather than other parts of the environment. Relatedly, there is work on using multiple agents in a single-agent setting to increase the diversity of the environment and hence the generality of the policy \shortcite{openendedlearningteamOpenEndedLearningLeads2021}, which we do cover.

Finally, we do not cover theoretical work on generalisation in RL\@. While there is recent work in this area \shortcite{duGoodRepresentationSufficient2020,malikWhenGeneralizableReinforcement2021} that is valuable, we focus on empirical research as it's more widely studied. 

\paragraph{Overview of the Survey.} The structure of the survey is as follows. We first briefly describe related work such as other surveys and overviews in \cref{section:relatedwork}. We introduce the formalism and terminology for ZSG in RL in  \cref{section:formalism}, including the relevant background. We then proceed to use this formalism to describe current benchmarks for ZSG in RL in \cref{section:benchmarks}, discussing both environments (\cref{subsection:environments}) and evaluation protocols (\cref{subsection:evalprotocol}). We categorise and describe work producing methods for tackling ZSG in \cref{section:methods}. Finally, we present a critical discussion of the current field, including recommendations for future work in both methods and benchmarks, in \cref{section:discussion}, and conclude with a summary of the key takeaways from the survey in \cref{section:conclusion}.

\paragraph{Contributions.} To summarise, our key contributions are:
\begin{itemize}
	\item We present a unified formalism and terminology for discussing the broad class of ZSG problems and breaking down the assumptions necessary to achieve ZSG, building on formalisms and terminology presented in multiple previous works \shortcite{ghoshWhyGeneralizationRL2021,hallakContextualMarkovDecision2015,harrisonADAPTZeroShotAdaptive2017,higginsDARLAImprovingZeroShot2018,songObservationalOverfittingReinforcement2019,perezGeneralizedHiddenParameter2020}. Our contribution here is the unification of these prior works into \emph{a clear formal description of the class of problems referred to as ZSG in RL}, which captures the full space of problems, which wasn't done by any one existing formalism.
	\item We propose a taxonomy of existing benchmarks that can be used to test for ZSG, splitting the discussion into categorising environments and evaluation protocols. Our formalism allows us to cleanly describe weaknesses of the purely Procedural Content Generation (PCG) approach to ZSG benchmarking and environment design: \emph{having a completely PCG environment limits the precision of the research that can be done on that environment}. We recommend that \emph{future environments should use a combination of PCG and controllable factors of variation}.
	\item \emph{We propose a categorisation of existing methods to tackle various ZSG problems}, motivated by a desire to make it easy both for practitioners to choose methods given a concrete problem and for researchers to understand the landscape of methods and where novel and useful contributions could be made. We point to many under-explored avenues for further research, including fast online adaptation, tackling RL-specific ZSG issues, novel architectures, model-based RL and environment instance generation.
	\item We critically discuss the current state of ZSG in RL research, recommending future research directions. In particular, we argue that \emph{building benchmarks would enable progress in offline RL generalisation and reward-function variation}, both of which are important settings. Further, we point to several different settings and evaluation metrics that are worth exploring: \emph{investigating context-efficiency and working in a continual RL setting} are both areas where future work is necessary.
\end{itemize}

\section{Related Work: Surveys In Reinforcement Learning Subfields}\label{section:relatedwork}

While there have been previous surveys of related subfields in RL, none have covered zero-shot generalisation in RL explicitly. \shortciteA{khetarpalContinualReinforcementLearning2020} motivated and surveyed continual reinforcement learning (CRL), which is closely related to ZSG in RL as both settings require adaptation to unseen tasks or environments; however, they explicitly do not discuss the zero-shot setting that is the concern of this paper (for more discussion of CRL see \cref{subsection:beyondsingle}). \shortciteA{chenOverviewRobustReinforcement2020} gave a brief overview of Robust RL (RRL) \shortcite{morimotoRobustReinforcementLearning2005}, a field aimed at tackling a specific form of environment model misspecification through worst-case optimisation. This is a sub-problem within the class of generalisation problems we discuss here, and \shortciteA{chenOverviewRobustReinforcement2020} only briefly survey the field. \shortciteA{albrechtAutonomousAgentsModelling2018} survey methods for modelling other agents, which can be seen as a form of generalisation problem (even in single-agent RL), if the environment contains a distribution over agents.
 
\shortciteA{zhaoSimtoRealTransferDeep2020} survey methods for sim-to-real transfer for deep RL in robotics. Sim-to-real is a concrete instantiation of the generalisation problem, and hence there is some overlap between our work and \shortciteA{zhaoSimtoRealTransferDeep2020}, but our work covers a much broader subject area, and some methods for sim-to-real transfer rely on data from the testing environment (reality), which we do not assume here. \shortciteA{muller-brockhausenProceduralContentGeneration2021} and \shortciteA{zhuTransferLearningDeep2021} survey methods for transfer learning in RL (TRL). TRL is related to generalisation in that both topics assume a policy is trained in a different setting to its deployment, but TRL generally assumes some form of extra training in the deployment or target environment, whereas we are focused on zero-shot generalisation. Finally, surveys on less related topics include \shortciteA{electronics9091363} who survey multi-task deep RL, \shortciteA{aminSurveyExplorationMethods2021} who survey exploration in RL, and \shortciteA{narvekarCurriculumLearningReinforcement2020} who survey curriculum learning in RL\@.

None of these surveys focuses on the zero-shot generalisation setting that is the focus of this work, and there is still a need for a formalism for the class of ZSG problems which will enable research in this field to discuss the differences between different problems.

\section{Formalising Zero-shot Generalisation In Reinforcement Learning}\label{section:formalism}

In this section, we present a formalism for understanding and discussing the class of zero-shot generalisation (ZSG) problems in RL\@. We first review the relevant background in supervised learning and RL before motivating the formalism itself. Formalising ZSG in this way shows that it refers to a \emph{class} of problems, rather than a specific problem, and hence research on ZSG needs to specify which group of ZSG problems it is tackling. Having laid out this class of problems in \cref{subsection:traintestcontext}, we discuss additional assumptions of structure that could make generalisation more tractable in \cref{subsection:addassump}; this is effectively specifying sub-problems of the wider ZSG problem.

\subsection{Background: Generalisation In Supervised Learning}\label{subsection:backgroundgen}

Generalisation in supervised learning is a widely studied area and hence is more mature than generalisation in RL (although it is still not well understood). In supervised learning, some predictor is trained on a training dataset, and the performance of the model is measured on a held-out testing dataset. It is often assumed that the data points in both the training and testing dataset are drawn independently and identically distributed (IID) from the same underlying distribution, although this is not  always the case (see for example the Domain Generalisation literature \shortcite{zhouDomainGeneralizationSurvey2022a}). Generalisation performance is then synonymous with the test-time performance, as the model needs to ``generalise'' to inputs it has not seen before during training. The generalisation gap in supervised learning for a model $\phi$ with training and testing data $D_{train}, D_{test}$ and loss function $\mathcal{L}$ is defined as
\begin{equation}\label{eq:slgengap}
  \textrm{GenGap}(\phi) := \mathbb{E}_{(x,y) \sim D_{test}} [\mathcal{L}(\phi, x, y)] - \mathbb{E}_{(x,y) \sim D_{train}} [\mathcal{L}(\phi,x,y)].
\end{equation}
This gap is normally used as a measure of generalisation specifically, independently of the train or test performance: for a given level of training performance, a smaller gap means a model generalises better. This metric isn't perfect, as a model that performs at random chance in both training and testing will get a gap of 0. Further, if the train and test datasets aren't drawn IID, then it's possible that the test dataset is easier (or harder), and hence a gap of zero doesn't necessarily imply perfect generalisation. However, it can be used to measure generalisation performance across benchmarks where absolute performance may not be comparable, or to motivate improvement from methods that may improve generalisation by lowering the gap without changing the test performance (in effect by lowering the training performance). These methods may then be combined with ones that improve training performance to improve the overall test performance, assuming that the methods don't conflict. We introduce this metric for completeness, as it has often been used in the literature in addition to test performance. In general, we think it's useful as a metric in addition to test performance, but not as a replacement for it. For more discussion, especially in the RL setting, see \cref{paragraph:genmetrics}.

One specific type of generalisation examined frequently in supervised learning which is relevant to RL is compositional generalisation \shortcite{hupkesCompositionalityDecomposedHow2020,keysersMEASURINGCOMPOSITIONALGENERALIZATION2020}. We explore a categorisation of compositional generalisation here introduced by \shortciteA{hupkesCompositionalityDecomposedHow2020}. While this was designed for generalisation in language, many of those forms are relevant for RL\@.
The five forms of compositional generalisation defined are:
\begin{enumerate}
	\item \textbf{systematicity}: generalisation via systematically recombining known parts and rules,
	\item \textbf{productivity}: the ability to extend predictions beyond the length seen in training data,
	\item \textbf{substitutivity}: generalisation via the ability to replace components with synonyms, 
	\item \textbf{localism}: if model composition operations are local vs.\ global,
	\item \textbf{overgeneralisation}: if models pay attention to or are robust to exceptions.
\end{enumerate}
For intuition, we will explore examples of some of these different types of compositional generalisation in a block-stacking environment. An example of \emph{systematicity} is the ability to stack blocks in new configurations once the basics of block-stacking are mastered. Similarly, \emph{productivity} can be measured by how many blocks the agent can generalise to, and the complexity of the stacking configurations. \emph{Substitutivity} can be evaluated by the agent's ability to generalise to blocks of new colours, understanding that the new colour does not affect the physics of the block. In \cref{paragraph:compositionalstructure} we discuss how assumptions of compositional structure in the RL environment can enable us to test these forms of generalisation.
In \cref{subsection:evalprotocol} we will discuss how some of these forms of generalisation can be evaluated in current RL benchmarks.

\subsection{Background: Reinforcement Learning}\label{subsection:backrl}

The standard formalism in RL is the Markov Decision Process (MDP). An MDP consists of a tuple $M = (S, A, R, T, p)$, where $S$ is the state space; $A$ is the action space; $R: S \times A \times S \rightarrow \R$ is the scalar reward function; $T(s'|s,a)$ is the possibly stochastic Markovian transition function; and $p(s_0)$ is the initial state distribution. We also consider partially observable MDPs (POMDPs). A POMDP consists of a tuple $M = (S, A, O, R, T, \phi, p)$, where $S, A, R, T$ and $p$ are as above, $O$ is the observation space, and $\phi : S \rightarrow O$ is the emission or observation function. In POMDPs, the policy only observes the observation of the state produced by $\phi$.

The standard problem in an MDP is to learn a policy $\pi(a|s)$ which produces a distribution over actions given a state, such that the cumulative reward of the policy in the MDP is maximised:
\[\pi^* = \underset{\pi \in \Pi}{\textrm{argmax }} \mathbb{E}_{s \sim p(s_0)} \left[\mathcal{R}(s)\right],\]
where $\pi^*$ is the optimal policy, $\Pi$ is the set of all policies, and $\mathcal{R}: S \to \R$ is the \emph{return} of a state, calculated as \[\mathcal{R}(s) := \mathbb{E}_{a_t\sim \pi(a_t|s_t), s_{t+1} \sim T(s_{t+1}|s_t, a_t)} \left[\sum_{t=0}^{\infty} R(s_t, a_t, s_{t+1})| s_0 = s \right].\] This is the total expected reward gained by the policy from a state $s$. The goal in a POMDP is the same, but with the policy taking observations rather than states as input. This sum may not exist if the MDP does not have a fixed horizon, so we normally use one of two other forms of the return, either assuming a fixed number of steps per episode (a \emph{horizon} $H$) or an exponential discounting of future rewards by a discount factor $\gamma$. Note that we formalise the policy here as Markovian (i.e.~that only takes the previous state as input) for simplicity, but the policy can take in the full history $(s_1, a_1, r_1, \ldots s_{t-1}, a_{t-1}, r_{t-1}, s_t)$ as input, for example using a recurrent neural network. We define the set of possible histories for a state and action space as $H[S,A] = \{(s_1, a_1, r_1, \dots s_{t-1}, a_{t-1}, r_{t-1}, s_t) | t \in \mathbb{N}\}$, similarly for an observation space. A policy being non-Markovian allows it to be adaptive (for further discussion see \cref{paragraph:0shottransfer}).

\subsection{Contextual Markov Decision Processes}\label{subsection:cmdp}

To talk about zero-shot generalisation, we desire a way of reasoning about a \emph{collection} of tasks, environment instances or levels: the need for generalisation emerges from the fact we train and test the policy on different collections of environment instances. Consider as a didactic example OpenAI Procgen \shortcite{cobbeLeveragingProceduralGeneration2020}: in this benchmark suite, each game is a collection of procedurally generated levels. Which level is generated is completely determined by a level seed, and the standard protocol is to train a policy on a fixed set of 200 levels and then evaluate performance on the full distribution of levels. Almost all other benchmarks share this structure: they have a collection of levels or tasks, which are specified by some seed, ID or parameter vector, and generalisation is measured by training and testing on different distributions over the collection of levels or tasks. To give a different example, in the Distracting Control Suite \shortcite{stoneDistractingControlSuite2021}, the parameter vector determines a range of possible visual distractions applied to the observation of a continuous control task, from changing the colours of objects to controlling the camera angle. While this set of parameter vectors has more structure than the set of seeds in Procgen, both can be understood within the framework we propose. See \cref{subsection:evalprotocol} for a discussion of the differences between these styles of environments.

To formalise the notion of a collection of tasks, we start with the Contextual Markov Decision Process (CMDP), as originally formalised by \shortciteA{hallakContextualMarkovDecision2015}, but using the alternative formalism from \shortciteA{ghoshWhyGeneralizationRL2021}. This formalism also builds on those presented by \shortciteA{doshi-velezHiddenParameterMarkov2013,perezGeneralizedHiddenParameter2020}, but we extend them to consider different distributions over context parameters; we include both observed and unobserved contexts settings; we include cases where the context is controllable; and we define formally how to produce subset CMDPs (see the end of this section for more discussion comparing the formalism we present here and existing works). 

\begin{definition}
A contextual MDP (CMDP) is a tuple 
\[\mathcal M = \left(S', A, O, R, T, C, \phi: S' \times C \rightarrow O, p(s'|c), p(c) \right).\]
$A, O, R, T, \phi$ are as in the definition of the POMDP in \cref{subsection:backrl}. $C$ is the context space (a set over which it is possible to have a distribution). The CMDP is a POMDP with state space $S := S' \times C$, initial state distribution $p((s',c)) = p(c)p(s'|c)$, that is the POMDP $\left(S' \times C, A, O, R, T, \phi, p(s'|c)p(c) \right)$. Hence, $R$ has type $R: S' \times C \rightarrow \R$ and $T((s,c), a)$ is the form of the transition probability distribution. For the tuple to be a CMDP, the transition function must be factored such that the context doesn't change within an episode, that is $T((s,c),a)((s',c')) = 0 \textrm{ if } c' \neq c$. We call $S'$ the underlying state space, and $p(c)$ the context distribution.
\end{definition}

To give an intuition for this definition, the context takes the role of the seed, ID or parameter vector which determines the level. Hence why it should not change within an episode, only between episodes. The CMDP is the entire collection of tasks or environment instances; in Procgen, each game (e.g.~starpilot, coinrun, etc.) is a separate CMDP\@. The context distribution $p(c)$ is what is used to determine the training and testing collections of levels, tasks or environment instances; in Procgen this distribution is uniform over the fixed 200 seeds at training time, and uniform over all seeds at testing time.

Note that this definition leaves it unspecified whether the context is observed by the agent: if $O = O' \times C$ for some underlying observation space $O'$ and $ \phi((s',c)) = (\phi'(s),c)$ for some underlying observation function $\phi': S' \rightarrow O'$ then we say the context is observed, otherwise it isn't. The context needs to be observed for the CMDP to be an MDP (and not a POMDP), but the opposite isn't true - even if the context is observed, $\phi'$ could not be the identity, in which case the POMDP isn't likely to be an MDP\@. Note that we will generally use ``MDP'' to refer to environments that are either MDPs or POMDPs.

As the reward function, transition function, initial state distribution and emission function all take the context as input, the choice of context determines everything about the resulting MDP apart from the action space, which we assume is fixed. Given a context $c^*$, we call the MDP resulting in the restriction of the CMDP $\mathcal{M}$ to the single context a \emph{context-MDP} $\mathcal{M}_{c^*}$. Formally, this is a new CMDP with $p(c) := 1 \textrm{ if } c = c^* \textrm{ else } 0$. This is a specific task or environment instance, for example, a single level of a game in Procgen, as specified by a single random seed that is the context.

Some MDPs have stochastic transition or reward functions. When these MDPs are simulated, researchers often have control of this stochasticity through the choice of a random seed. In theory, these stochastic MDPs could be considered deterministic contextual MDPs, where the context is the random seed. We do not consider stochastic MDPs as automatically contextual in this way and assume that the random seed is always chosen randomly, rather than being modelled as a context. This more closely maps to real-world scenarios with stochastic dynamics where we cannot control the stochasticity.

\subsection{Training And Testing Contexts}\label{subsection:traintestcontext}

We now describe the class of generalisation problems we focus on, using the CMDP formalism. As mentioned, the need for generalisation emerges from a difference between the training and testing environment instances, and so we want to specify both a set of training context-MDPs and a testing set. We specify these sets of context-MDPs by their context sets, as the context uniquely determines the MDP\@.

First, we need to describe how to use training and testing context sets to create new CMDPs.

\begin{definition}\label{def:cmdpsubset}
For any CMDP $\mathcal{M} = \left(S', A, O, R, T, C, \phi, p(s'|c), p(c) \right)$, we can choose a subset of the context set $C' \subseteq C$, and then produce a new CMDP
	\[\mathcal{M}|_{C'} = \left(S', A, O, R, T, C', \phi, p(s'|c), p'(c) \right)\]
where $p'(c) = \frac{p(c)}{Z} \textrm{ if } c \in C' \textrm{ else } 0$ and $Z$ is a renormalisation term $Z = \sum_{c \in C'}p(c)$ that ensures $p'(c)$ is a probability distribution.
\end{definition}

This allows us to split the total collection of context-MDPs into smaller subsets, as determined by the contexts. For example, in Procgen any possible subset of the set of all seeds can be used to define a different version of the game with a limited set of levels.

For the objective, we use the expected return of a policy:
\begin{definition}\label{defn:expectedreturn}
	For any CMDP $\mathcal{M}$ we can define the expected return of a policy in that CMDP as
	\[\textbf{R}(\pi, \mathcal{M}) := \mathbb{E}_{c \sim p(c)}[\mathcal{R}(\pi, \mathcal{M}_c)],\]
	where $\mathcal{R}$ is the expected return of a policy in a (context) MDP and $p(c)$ is the context distribution as before.
\end{definition}

We can now formally define the Zero-Shot Policy Transfer (ZSPT) problem class.

\begin{definition}[Zero Shot Policy Transfer]
	A ZSPT problem is defined by a choice of CMDP $\mathcal{M}$ with context set $C$ and a choice of training and testing context sets $C_{\textrm{train}}, C_{\textrm{train}} \subseteq C$. The objective is to produce a non-Markovian policy $\pi: H[O,A] \rightarrow A$ which maximises the expected return in the testing CMDP $\mathcal{M}|_{C\textrm{test}}$:
	\[J(\pi) := \textbf{R}(\pi, \mathcal{M}|_{C\textrm{test}}).\]
	This policy can be produced through interaction with the training CMDP $\mathcal{M}|_{C_\textrm{train}}$ for a fixed number of environment and episode samples $N_s, N_e$ respectively.
\end{definition}\label{def:zspt}

ZSG research is generally concerned with developing algorithms that can solve a variety of ZSPT problems. For example, in Procgen we aim to produce an algorithm that can solve the ZSPT problem for every game. Specifically, we want to achieve the highest return possible on the testing distribution (which is the full distribution over levels) after training for 25 million steps ($N_s = 25 \times 10^6, N_e = \infty$) on the training distribution of levels (which is a fixed set of 200 levels). The name \emph{Zero-Shot Policy Transfer} comes from prior works \shortcite{harrisonADAPTZeroShotAdaptive2017,higginsDARLAImprovingZeroShot2018}.

Some algorithms assume that the context distribution can be adjusted during interaction with the training CMDP, as long as sampled contexts are only within the fixed training context set:

\begin{definition}[ZSPT controllable context]
		A \emph{controllable context} ZSPT problem is the same as a ZSPT problem above, except the learning algorithm can adjust the context distribution of the training CMDP $p_{\textrm{train}}(c)$ during training, so long as it maintains the property of only sampling from the training context set: $p_{\textrm{train}}(c) = 0 \textrm{ if } c \not\in C_{\textrm{train}}$
\end{definition}\label{def:zsptcc}

\emph{Note that this formalism defines a \emph{class} of problems, each determined by a choice of CMDP, training and testing context sets and whether the context is controllable}. This means that we do not make any assumptions about shared structure within the CMDP between context-MDPs: for any specific problem some assumption of this kind (either implicit or explicit) will likely be required for learning to occur (\cref{subsection:addassump}), but we do not believe there is a unifying assumption behind all ZSG problems apart from those stated here.

\paragraph{Evaluating Zero-Shot Generalisation.}\label{paragraph:genmetrics}
As in supervised learning, we can consider the gap between training and testing performance as a measure of generalisation. We define this analogously to how it's defined in supervised learning (\cref{eq:slgengap}), swapping the ordering between training and testing (as we maximise reward, rather than minimise loss): 

\begin{equation}\label{eq:rlgengap}
  \textrm{GenGap}(\pi) := \textbf{R}(\pi, \mathcal{M}|_{C_{\textrm{train}}}) - \textbf{R}(\pi, \mathcal{M}|_{C_{\textrm{test}}}).
\end{equation}

This metric is frequently used in the literature in addition to test performance to evaluate ZSG algorithms \shortcite{jiangPrioritizedLevelReplay2021,raileanuAutomaticDataAugmentation2021,raileanuDecouplingValuePolicy2021}. In general, it's not clear what the best evaluation metric for ZSG algorithms is. In supervised learning, the generalisation capabilities of different algorithms are usually evaluated via final performance on an evaluation task. When the tasks used to evaluate a model are close to (or the same as) the tasks that the model will eventually be deployed on, it is clear that final performance is a good metric to evaluate on. However, in RL the benchmark tasks we use are often very dissimilar to the eventual real-world tasks we want to apply these algorithms to. Further, RL algorithms are currently still quite brittle and performance can vary greatly depending on hyperparameter tuning and the specific task being used \shortcite{hendersonDeepReinforcementLearning2019}. In this setting, we may care more about the zero-shot generalisation \emph{potential} of algorithms by decoupling generalisation from training performance and evaluating using the generalisation gap instead. For example, if algorithm A has higher testing performance than algorithm B, but A also has a much larger generalisation gap, we may prefer to use algorithm B in a new setting, so we have better assurance that the deployment performance won't deviate as much from the training performance, and the algorithm may be more robust. This is the reason the previous literature has often reported this metric alongside test performance.

However, the generalisation gap in RL has the same problems as discussed for the supervised learning generalisation gap: a gap of zero doesn't necessarily imply good performance (i.e.~a random policy is likely to get a gap of 0), and if the reward functions aren't comparable across training and testing, then the magnitude of the gap my not be informative (and it could then only be used to compare between different algorithms). This means using it as the only metric for improved performance will likely not lead to robust progress in ZSG\@. Further, given how broad the current set of assumptions is, it is unlikely there is a single general measure of progress towards tackling ZSG\@: across such a broad problem class, objectives may even be conflicting \shortcite{wolpertNoFreeLunch1997}.

Therefore, our recommendation is first and foremost to focus on problem-specific benchmarks and revert to the SL standard of using overall performance in specific settings (e.g.~visual distractors, stochastic dynamics, sparse reward, hard exploration). The generalisation performance of various RL algorithms is likely contingent on the type of environment they are deployed on and therefore careful categorisation of the type of challenges present at deployment is needed to properly evaluate ZSG capability (for further discussion see \cref{paragraph:procgengen} and \cref{subsection:rwrlg}). As in the literature, generalisation gap can be used as an additional auxiliary metric to evaluate the performance of ZSG algorithms as well as test performance, to either break ties between algorithms with very similar test time performance, or to inform users in situations where it is more important to have strong assurances on test time performance than to have expected test time performance as high as possible.

\subsection{Real World Examples of This Formalism}

We chose this formalism as it is simple to understand, captures all the problems we are interested in, and is based on prior work. To further justify why this formalism is useful, and to give intuition about how it can be used in a variety of settings, we give several examples of real-world scenarios where this formalism naturally applies:
\begin{itemize}
	\item Sim-to-real is a classic problem of ZSG and one which can be captured in this framework. Here the outer CMDP is a union between the simulation and real-world MDPs. The context set will be split into those contexts which correspond to simulation, and those that correspond to reality. The context generally conditions the dynamics, observation function and state distribution, but likely not the reward (so $\forall s',c:R((s',c)) = R'(s')$). Domain randomisation approaches are motivated by the idea that producing a wide range of possible contexts in simulation (the training CMDP) will make it more likely that the testing distribution of contexts is closer to the expanded training distribution. In a simulation setting, we'd normally assume access to the context distribution, and that the context could be made observable, but that the context won't be observable at testing time, and so it may not be useful to have a policy that explicitly conditions on the context.
	\item Healthcare is a promising domain for deploying future RL methods, as there are many sequential decision-making problems. For example, the task of diagnosing and treating individual patients can be understood as a CMDP where the patient effectively specifies the context: patients will react differently to tests and treatments (dynamics variation) and may provide different measurements (state variation). Generalising to treating new patients is then exactly generalising to novel contexts. In this setting, we may be able to assume some part of the context (or some information about the context) is observable, as we will have access to the patient's medical history and personal information.
	\item Autonomous vehicles are another area where RL methods could be applied. These vehicles will be goal-conditioned in some sense, such that they can perform different journeys, which means that the context will likely control the reward function and that the part of the context that controls the read function will be observable (so the policy knows what task to perform). Driving in different locations (different contexts changing the initial state distribution), under different weather and lighting conditions due to the time of day (observation functions) and on different road surfaces (transition functions) are all problems that need to be tackled by these systems. We can understand this in the CMDP framework, where the context contains information about the weather, time of day, location and goal, as well as information about the state of the current vehicle. Some of this context will be observed directly, and some may be inferred from observation. In this setting, we may only be able to train in certain contexts (i.e.~certain cities, or restricted weather conditions), but we require the policy to generalise zero-shot to the unseen contexts well.
\end{itemize}

\subsection{Additional Assumptions For More Feasible Generalisation}\label{subsection:addassump}
In choosing the CMDP formalism we opted to formalise ZSG in a way that captures the full class of problems we are concerned with, but this means that it is almost certainly impossible to prove any formal theoretical guarantees on learning performance using solely the CMDP structural assumptions. While we do not prove this, it is easy to see how one could design pathological CMDPs where generalisation to new contexts is entirely impossible without strong domain knowledge of the new contexts.

To have any chance of solving a specific ZSG problem then, further assumptions (either explicit or implicit) have to be made. These could be assumptions on the type of variation, the distributions from which the training and testing context sets are drawn, or additional underlying structure in the context set. We describe several popular or promising assumptions here and note that the taxonomy in \cref{section:benchmarks} also acts as a set of possible additional assumptions to make when tackling a ZSG problem.

\paragraph{Assumptions on the Training and Testing Context Set Distributions.}

One assumption which is often made is that while the training and testing context sets are not identical, the elements of the two sets have been drawn from the same underlying distribution, analogously to the IID data assumption in supervised learning. For example, this is the setup of OpenAI Procgen~\shortcite{cobbeLeveragingProceduralGeneration2020}, where the training context set is a set of 200 seeds sampled uniformly at random from the full distribution of seeds, and the full distribution is used as the testing context set.

However, many works on ZSG in RL do not assume that the train and test environment instances are drawn from the same distribution. This is often called \emph{Domain Generalisation}, where we refer to the training and testing environments instances as different \emph{domains} that may be similar but are not from the same underlying generative distribution. Concrete examples occur in robotics such as the \emph{sim-to-real} problem.

Note that while the testing context set could be a single context, this would likely lead to a not particularly robust algorithm - it's possible that it overfits to the problem of producing a policy that performs well on this specific context, which may not generalise to other similar contexts (which is the high-level goal of this research direction).

\paragraph{Further Formal Assumptions of Structure.}
Another kind of assumption that can be made is on the structure of the CMDP itself, e.g.~the context space or transition function. There are several families of MDPs with additional structure which could enable ZSG\@. However, these assumptions are often not explicitly made when designing benchmarks and methods, which can make understanding why and how generalisation occurs difficult. A detailed discussion and formal definitions for these structures can be found in \cref{app:structure}, but we provide a high-level overview here, focusing on assumptions that have been used in practice, and those that hold particular promise for ZSG\@.

An example of a structured MDP that has been used to improve generalisation is the \emph{block MDP} \shortcite{duProvablyEfficientRL2019}. It assumes a \emph{block structure} in the mapping from a latent state space to the given observation space, or that there exists another MDP described by a  smaller state space with the same behaviour as the given MDP\@. This assumption is relevant in settings where we only have access to high-dimensional, unstructured inputs, but know that there exists a lower-dimensional state space that gives rise to an equivalent MDP\@. \shortciteA{duProvablyEfficientRL2019} use this assumption for improved bounds on exploration that relies on the size of the latent state space rather than the given observation space. \shortciteA{zhangInvariantCausalPrediction2020} develop a representation learning method that disentangles relevant from irrelevant features, improving generalisation to environments instances where only the irrelevant features change, a simple form of \emph{systematicity} (\cref{subsection:backgroundgen}, \shortciteR{hupkesCompositionalityDecomposedHow2020}). This is a rare example of a method explicitly utilising additional assumptions of structure to improve generalisation. Block MDPs can be combined with contextual MDPs by introducing an emission mapping from state space to observation space that is also dependent on context, as defined by \shortciteA{sodhaniBlockContextualMDPs2022}. 

Factored MDPs~\shortcite{boutilierStochasticDynamicProgramming2000,strehlEfficientStructureLearning2007} can be used to describe object-oriented environments or multi-agent settings where the state space can be broken up into independent factors, i.e.~with sparse relationships over the one-step dynamics. This can be leveraged to learn dynamics models that explicitly ignore irrelevant factors in prediction or to compute improved sample complexity bounds for policy learning \shortcite{haoOnlineSparseReinforcement2021} and seems particularly relevant for generalisation as additional structure in the context set could map onto the factored structure in the transition and reward functions. An initial example of using a similar formalism to a factored MDP in a multi-domain RL setting is demonstrated by \shortciteA{huangAdaRLWhatWhere2021a}, although it does not target the zero-shot policy transfer setting directly. 
We note that contextual MDPs can be trivially represented by factored MDPs with two factors, the state and context. However, factored MDPs are capable of modelling more structure in a domain if present. Therefore, if an environment is capable of being modelled by a factored MDP in addition to a contextual MDP, better generalisation guarantees and results are likely possible if this structure is exploited.
We hope to see more work applying these kinds of structural assumptions to the zero-shot generalisation problems discussed in this work.

\subsection{Remarks And Discussion}\label{subsection:formdiscussion}
\paragraph{Angles to Tackle the ZSPT Problem.}
While we aim to improve test-time performance \cref{defn:expectedreturn}, we often do not have access to that performance metric directly (or will not when applying our methods in the real world). In research, to develop algorithms that improve the test-time performance, we could aim to produce algorithms that (when compared to existing work) either (1) increase the train-time performance while keeping the generalisation gap constant; (2) decrease the generalisation gap while keeping the train-time reward constant; or (3) do a mixture of the two approaches. Work in RL not concerned with generalisation tends to (implicitly) take the first approach, assuming that the generalisation gap will not change.\footnote{If the training environment instances are identical to the testing environment instances, then the generalisation gap will always be 0.} Work on ZSG in RL instead normally aims at (2) reducing the generalisation gap explicitly, which may reduce train-time performance but increase test-time performance. Some work also aims at (1) improving train-time performance in a way that is likely to keep the generalisation gap constant.

\paragraph{Motivating Zero-Shot Policy Transfer}\label{paragraph:0shottransfer}

In this work, we focus on zero-shot policy transfer \shortcite{harrisonADAPTZeroShotAdaptive2017,higginsDARLAImprovingZeroShot2018}: a policy is learned from the training CMDP and evaluated zero-shot in the testing CMDP\@. This field is important for several reasons.

First, as mentioned above, current deep RL algorithms often aren't safe or sample-efficient enough to perform online learning in the real world. Hence, a significant amount of offline or in-simulation training is required, and policies need to generalise zero-shot to the real-world deployment setting. As access to more compute and richer simulations becomes available, we expect many successful real-world deployments of RL to follow this workflow at least in part: training offline or in simulation and then transferring the policy zero-shot to the deployment environment. Even if the policy will continue learning during deployment, it still needs to be reasonably good at deployment time (i.e.~zero-shot), otherwise it wouldn't be safe to deploy it. In this way, we view work on ZSG as mostly complementary to work on continual RL, as we think both are important for enabling the deployment of robust and competent RL policies. Note that there may be tradeoffs between good zero-shot performance and good continual learning performance, and how to choose between these two desiderata will be determined by the specific problem setting being faced.

Second, from a safety, interpretability and verification perspective, ZSG may be preferable to a continually updated policy. It's likely in high-stakes scenarios that models will be verified \shortcite{katzMarabouFrameworkVerification2019} or audited with interpretability or explainability methods \shortcite{milaniSurveyExplainableReinforcement2022}, and that this process will be expensive. In this scenario, it will be beneficial to have a single model which performs well zero-shot without having to be continually updated, as after each update these verification and auditing steps will likely have to be repeated.

Finally, note that while we do not cover methods that relax the zero-shot assumption, we believe that in a real-world scenario it will likely be possible to do so.\footnote{For example by using unsupervised data or samples in the testing environment instances, utilising some description of the contexts such that zero-shot generalisation is possible, or enabling the agent to train in an online way in the testing context-MDPs.} However, zero-shot policy transfer is still a useful problem to tackle, as solutions are likely to help with a wide range of settings resulting from different relaxations of the assumptions made here: zero-shot policy transfer algorithms can be used as a base which is then built upon with domain-specific knowledge and extra data.

Note that while ``training'' and ``learning'' are contentious terms, our definition grounds them in the production of a non-Markovian policy after some number of samples from the training context MDP\@. The fact that the objective is the expected testing return within a \emph{single episode} of the non-Markovian policy means that online learning or adaptation across more than 1 episode isn't possible, and so the adaptation would have to happen within a single episode to be useful (i.e.~using a recurrent policy). Several methods do take this approach, as described in \cref{subsubsection:adaptingonline}. To be clear, we are grounding ourselves in this specific \emph{objective}, and not placing any restrictions on the \emph{properties} of methods as long as they satisfy the constraints: policies need not be Markovian, and can adapt within a single episode to the environment instances they're placed in if that improves performance.

\paragraph{Relationship to Previous Notions of Generalisation in RL.}
Historically, generalisation in RL has referred to the notion of generalising to novel states or state-action pairs within a single MDP\@. While this notion is useful in the online learning single-MDP setting that is the focus of that work, we here focus on a more recent and more realistic setting where the policy is trained on a collection of MDPs and then deployed on possibly unseen MDPs. This setting more closely mirrors the supervised learning notion of generalisation. We believe this type of workflow for deploying RL-trained policies in the real world is much more feasible than training a policy online from scratch, as current RL algorithms aren't sample-efficient or safe enough to train online in the real world. This means a large amount of off-line or in-simulation training will have to occur before the policy is deployed, and the deployed policy will still need to perform reasonably well as soon as it's deployed (i.e.~zero-shot).

\paragraph{Relationship to Previous Formalisms of Collections or Distributions of MDPs.}
As mentioned previously, the formalism we presented above builds on multiple previous works. Here we briefly present the differences between our formalism and these works.

\shortciteA{doshi-velezHiddenParameterMarkov2013} present Hidden-parameter MDPs (Hi-MDPs). These are MDPs with a hidden parameter which controls the transition function and a distribution over the set of these hidden parameters that implicitly defines a set of MDPs. \shortciteA{perezGeneralizedHiddenParameter2020} builds on Hi-MDPs to present Generalised Hidden-parameter MDPs (GHP-MDPs), where the hidden parameters now also control the reward function in addition to the dynamics. Our work uses a context parameter which is analogous to the hidden parameter, which may be hidden or observed, and that controls the initial state distribution, transition function and reward function, rather than just the transition and reward functions. We also more formally discuss how different distributions of context parameters may be used during training and testing.

\shortciteA{harrisonADAPTZeroShotAdaptive2017,higginsDARLAImprovingZeroShot2018} introduce variations on the term "zero-shot policy transfer", which we use as the name for the formal class of problems we study. They both study problems within this class but don't formalise the entire class of problems as we do here, instead focusing on presenting methods for improving performance in the empirical settings they investigate.

\shortciteA{songObservationalOverfittingReinforcement2019} discusses a distribution over MDPs which are sampled from during training, and uses a kind of CMDP but where the context parameter adjusts only the observation function, rather than any other parts of the MDP. They also assume the distribution over MDPs is the same between training and testing.

\shortciteA{hallakContextualMarkovDecision2015,ghoshWhyGeneralizationRL2021} present Contextual MDPs (CMDPs), which is the formalism we use as the base of our definitions. \shortciteA{ghoshWhyGeneralizationRL2021} present a formalism based on the context being part of an underlying state space, which is the one we use, while \shortciteA{hallakContextualMarkovDecision2015} present a formalism more similar to Hi-MDPs, where there is a collection of MDPs parameterised by a context variable. Both of these works don't consider a shift in distribution over contexts between training and testing apart from a shift to the full distribution in \shortciteA{ghoshWhyGeneralizationRL2021}. They don't present a formal definition of the policy class and don't define problem settings where the context distribution is controllable.

\section{Benchmarks For Zero-shot Generalisation In Reinforcement Learning}\label{section:benchmarks}

In this section, we give a taxonomy of benchmarks for ZSG in RL\@. A key split in the factors of variation for a benchmark is those factors concerned with the environment and those concerned with the evaluation protocol. A benchmark task is a combination of a choice of the environment (a CMDP, covered in \cref{subsection:environments}) and a suitable evaluation protocol (a train and test context set, covered in \cref{subsection:evalprotocol}). This means that all environments support multiple possible evaluation protocols, as determined by their context sets.

Having categorised the set of benchmarks, we point out the limitations of the purely PCG approach to building environments (\cref{paragraph:procgengen}), as well as discuss the range of difficulty among ZSG problems (\cref{paragraph:genexpect}). More discussion of future work on benchmarks for ZSG can be found in \cref{subsection:beyondsingle,subsection:rwrlg,subsection:strongervariation}.

\begin{table}[ht]
	\ssmall
\centering
\begin{tabularx}{\linewidth}{Xlll}
\toprule
\textbf{Name} & \textbf{Style} & \textbf{Contexts} & \textbf{Variation} \\ \midrule\rowcolor{Gray}
\textit{Alchemy}              \textdagger{} (\shortciteauthor{wangAlchemyStructuredTask2021})                                                 & 3D             & PCG                    & \textcolor{CBGreen}{D}, \textcolor{CBOrange}{R}, \textcolor{CBPink}{S}                         \\
\textit{Animal-AI}            (\shortciteauthor{crosbyAnimalAITestbedCompetition2020})                                                        & 3D             & D-C, D-O           & \textcolor{CBPink}{S}, \textcolor{CBBlue}{O}                         \\\rowcolor{Gray}
\textit{Atari Game Modes}     (\shortciteauthor{machadoRevisitingArcadeLearning2017})                                                         & Arcade         & D-C                  & \textcolor{CBGreen}{D}, \textcolor{CBBlue}{O}, \textcolor{CBPink}{S}                       \\                
\textit{BabyAI}               (\shortciteauthor{chevalier-boisvertBabyAIPlatformStudy2019})                                                   & Grid, LC       & D-C, D-O, PCG      & \textcolor{CBOrange}{R}, \textcolor{CBPink}{S}                                              \\\rowcolor{Gray}
\textit{CARL}                 (\shortciteauthor{benjaminsCARLBenchmarkContextual2021})                                                        & Varied         & Con, D-C, D-O      & \textcolor{CBGreen}{D}, \textcolor{CBBlue}{O}, \textcolor{CBOrange}{R}, \textcolor{CBPink}{S}   \\
\textit{CARLA}                (\shortciteauthor{fanSECANTSelfExpertCloning2021,zhuRobosuiteModularSimulation2020})                           & 3D, Driving    & D-C                  & \textcolor{CBBlue}{O}                                                               \\\rowcolor{Gray}
\textit{CausalWorld}          \textdagger{} (\shortciteauthor{ahmedCausalWorldRoboticManipulation2020})                                       & 3D, ConCon     & Con, D-C, D-O      & \textcolor{CBGreen}{D}, \textcolor{CBBlue}{O}, \textcolor{CBOrange}{R}, \textcolor{CBPink}{S}   \\
\textit{Construction}         (\shortciteauthor{bapstStructuredAgentsPhysical2019})																														& 2D, Structured & Con, D-C, D-O, PCG & \textcolor{CBOrange}{R}, \textcolor{CBPink}{S}   \\                     \rowcolor{Gray}
\textit{Crafter}              (\shortciteauthor{hafnerBenchmarkingSpectrumAgent2021})                                                         & Arcade, Grid   & PCG                    & \textcolor{CBPink}{S}                                                                  \\
\textit{Crafting gridworld}   (\shortciteauthor{chenAskYourHumans2021})                                                                       & Grid, LC       & D-C                  & \textcolor{CBOrange}{R}, \textcolor{CBPink}{S}                                              \\\rowcolor{Gray}
\textit{DACBench}             (\shortciteauthor{eimerDACBenchBenchmarkLibrary2021})																														& Structured     & PCG, D-C							& \textcolor{CBGreen}{D}, \textcolor{CBOrange}{R}, \textcolor{CBPink}{S}   \\                     
\textit{DCS}                  (\shortciteauthor{stoneDistractingControlSuite2021})                                                            & ConCon         & Con, D-C             & \textcolor{CBBlue}{O}                                                               \\\rowcolor{Gray}
\textit{DistractingCarRacing} (\shortciteauthor{tangNeuroevolutionSelfInterpretableAgents2020,brockmanOpenAIGym2016})												& Arcade         & D-C                  & \textcolor{CBBlue}{O}                                                               \\                
\textit{DistractingVizDoom}   (\shortciteauthor{tangNeuroevolutionSelfInterpretableAgents2020,openaiOpenAIGymDoomTakeCoverv02016})                     & 3D             & D-C                  & \textcolor{CBBlue}{O}                                                               \\\rowcolor{Gray}
\textit{DMC-GB}               (\shortciteauthor{hansenGeneralizationReinforcementLearning2021})                                               & ConCon         & Con, D-C             & \textcolor{CBBlue}{O}                                                               \\                
\textit{DMC-Remastered}       (\shortciteauthor{grigsbyMeasuringVisualGeneralization2020})                                                    & ConCon         & Con, D-C             & \textcolor{CBBlue}{O}                                                               \\\rowcolor{Gray}
\textit{DM-Memory}						(\shortciteauthor{fortunatoGeneralizationReinforcementLearners2020})                                            & Arcade, 3D     & PCG, Con, D-O, D-C & \textcolor{CBOrange}{R}, \textcolor{CBGreen}{D}, \textcolor{CBPink}{S}              \\
\textit{GenAsses}             (\shortciteauthor{packerAssessingGeneralizationDeep2019})                                                       & ConCon         & Con                    & \textcolor{CBGreen}{D}, \textcolor{CBPink}{S}                                             \\\rowcolor{Gray}
\textit{GVGAI}                (\shortciteauthor{perez-liebanaGeneralVideoGame2019})                                                           & Grid           & D-C                  & \textcolor{CBGreen}{D}, \textcolor{CBBlue}{O}, \textcolor{CBPink}{S}                       \\
\textit{HALMA}                \textdagger{} (\shortciteauthor{xieHALMAHumanlikeAbstraction2021})                                              & Grid           & D-C, D-O           & \textcolor{CBBlue}{O}, \textcolor{CBPink}{S}                       \\\rowcolor{Gray}
\textit{iGibson}              (\shortciteauthor{fanSECANTSelfExpertCloning2021,shenIGibsonSimulationEnvironment2021})                        & 3D             & D-C                  & \textcolor{CBBlue}{O}, \textcolor{CBPink}{S}                                            \\                
\textit{Jericho}              \textdagger{} (\shortciteauthor{hausknechtInteractiveFictionGames2020})                                         & Text           & D-C                  & \textcolor{CBGreen}{D}, \textcolor{CBOrange}{R}, \textcolor{CBPink}{S}                         \\\rowcolor{Gray}
\textit{JumpingFromPixels}    (\shortciteauthor{tachetLearningInvariancesPolicy2020})                                                         & Arcade         & Con                    & \textcolor{CBPink}{S}                                                                  \\                
\textit{KitchenShift}          (\shortciteauthor{xingKitchenShiftEvaluatingZeroShot2021})                                                      & 3D, ConCon     & D-C                  & \textcolor{CBBlue}{O}, \textcolor{CBPink}{S}, \textcolor{CBOrange}{R}                                            \\\rowcolor{Gray}
\textit{Malmo}          (\shortciteauthor{johnsonMalmoPlatformArtificial2016})																																& 3D, Arcade     & D-C, D-O           & \textcolor{CBOrange}{R}, \textcolor{CBPink}{S}                                            \\
\textit{MarsExplorer}         (\shortciteauthor{koutrasMarsExplorerExplorationUnknown2021})                                                   & Grid           & PCG                    & \textcolor{CBPink}{S}                                                                  \\\rowcolor{Gray}
\textit{MazeExplore}          (\shortciteauthor{harriesMazeExplorerCustomisable3D2019})                                                       & 3D             & PCG                    & \textcolor{CBBlue}{O}, \textcolor{CBPink}{S}                                            \\
\textit{MDP Playground}       \textdagger{} (\shortciteauthor{rajanMDPPlaygroundDesign2021})                                                  & ConCon, Grid   & Con, D-C, D-O      & \textcolor{CBGreen}{D}, \textcolor{CBBlue}{O}, \textcolor{CBOrange}{R}, \textcolor{CBPink}{S}   \\\rowcolor{Gray}
\textit{Meta-World}           \textdagger{} (\shortciteauthor{yuMetaworldBenchmarkEvaluation2019})                                            & 3D, ConCon     & Con, D-C             & \textcolor{CBOrange}{R}, \textcolor{CBPink}{S}                                              \\
\textit{MetaDrive}            (\shortciteauthor{liMetaDriveComposingDiverse2021})                                                             & 3D, Driving    & D-C, D-O, PCG      & \textcolor{CBGreen}{D}, \textcolor{CBPink}{S}                                             \\\rowcolor{Gray}
\textit{MiniGrid}             (\shortciteauthor{chevalier-boisvertMinimalisticGridworldEnvironment2021})                                      & Grid           & PCG                    & \textcolor{CBPink}{S}                                                                  \\
\textit{MiniHack}             (\shortciteauthor{samvelyanMiniHackPlanetSandbox2021})                                                          & Grid           & D-C, D-O, PCG      & \textcolor{CBPink}{S}                                                                  \\\rowcolor{Gray}
\textit{NaturalEnvs CV}       (\shortciteauthor{zhangNaturalEnvironmentBenchmarks2018})                                                       & Grid           & PCG                    & \textcolor{CBBlue}{O}, \textcolor{CBOrange}{R}, \textcolor{CBPink}{S}                        \\
\textit{NaturalEnvs MuJoCo}   (\shortciteauthor{zhangNaturalEnvironmentBenchmarks2018})                                                       & ConCon         & D-C                  & \textcolor{CBBlue}{O}                                                               \\\rowcolor{Gray}
\textit{NLE}                  (\shortciteauthor{kuttlerNetHackLearningEnvironment2020})                                                       & Grid           & PCG                    & \textcolor{CBPink}{S}                                                                  \\
\textit{Noisy MuJoCo}         (\shortciteauthor{zhaoInvestigatingGeneralisationContinuous2019})                                               & ConCon         & Con, D-C             & \textcolor{CBGreen}{D}, \textcolor{CBBlue}{O}                                          \\\rowcolor{Gray}
\textit{NovelGridworlds}      (\shortciteauthor{goelnovelgridworlds})                                                                         & Grid           & D-C                  & \textcolor{CBGreen}{D}, \textcolor{CBPink}{S}                                             \\
\textit{Obstacle Tower}       (\shortciteauthor{julianiObstacleTowerGeneralization2019})                                                      & 3D             & D-C, PCG             & \textcolor{CBBlue}{O}, \textcolor{CBPink}{S}                                            \\\rowcolor{Gray}
\textit{OffRoadBenchmark}       (\shortciteauthor{dosovitskiyCARLAOpenUrban2017,hanNewOpenSourceOffRoad2021})                                 & 3D, Driving    & D-C,                 & \textcolor{CBBlue}{O}, \textcolor{CBPink}{S}, \textcolor{CBOrange}{R}                       \\
\textit{OpenAI Procgen}       (\shortciteauthor{cobbeLeveragingProceduralGeneration2020})                                                     & Arcade         & PCG                    & \textcolor{CBBlue}{O}, \textcolor{CBPink}{S}                                            \\\rowcolor{Gray}
\textit{OverParam Gym}        (\shortciteauthor{songObservationalOverfittingReinforcement2019})                                               & ConCon         & Con                    & \textcolor{CBBlue}{O}                                                               \\
\textit{OverParam LQR}        (\shortciteauthor{songObservationalOverfittingReinforcement2019})                                               & LQR            & Con                    & \textcolor{CBBlue}{O}                                                               \\\rowcolor{Gray}
\textit{ParamGen}             (\shortciteauthor{keParametricGeneralizationBenchmarking2021})                                                  & 3D, LC         & D-C, D-O           & \textcolor{CBOrange}{R}, \textcolor{CBPink}{S}                                              \\
\textit{RLBench}              \textdagger{} (\shortciteauthor{jamesRLBenchRobotLearning2019})                                                 & 3D, ConCon, LC & Con, D-C, D-O      & \textcolor{CBOrange}{R}, \textcolor{CBPink}{S}                                              \\\rowcolor{Gray}
\textit{RoboSuite}            (\shortciteauthor{fanSECANTSelfExpertCloning2021,zhuRobosuiteModularSimulation2020})                           & 3D, ConCon     & D-C                  & \textcolor{CBBlue}{O}                                                               \\
\textit{Rogue-gym}            (\shortciteauthor{kanagawaRogueGymNewChallenge2019})                                                            & Grid           & PCG                    & \textcolor{CBPink}{S}                                                                  \\\rowcolor{Gray}
\textit{RTFM}                 (\shortciteauthor{zhongRTFMGeneralisingNovel2021})                                                              & Grid, LC       & PCG                    & \textcolor{CBGreen}{D}, \textcolor{CBOrange}{R}, \textcolor{CBPink}{S}                         \\
\textit{RWRL}                 \textdagger{} (\shortciteauthor{dulac-arnoldEmpiricalInvestigationChallenges2021})                              & ConCon         & Con                    & \textcolor{CBGreen}{D}                                                                \\\rowcolor{Gray}
\textit{Sokoban}              (\shortciteauthor{weberImaginationAugmentedAgentsDeep2018})                                                     & Grid           & PCG                  & \textcolor{CBPink}{S}                                                                \\
\textit{TextWorld}            \textdagger{} (\shortciteauthor{coteTextWorldLearningEnvironment2019})                                          & Text           & Con, D-C, PCG        & \textcolor{CBGreen}{D}, \textcolor{CBBlue}{O}, \textcolor{CBOrange}{R}, \textcolor{CBPink}{S}   \\\rowcolor{Gray}
\textit{Toybox}               \textdagger{} (\shortciteauthor{toschToyboxSuiteEnvironments2019})                                              & Arcade         & Con, D-C, D-O      & \textcolor{CBGreen}{D}, \textcolor{CBBlue}{O}, \textcolor{CBPink}{S}                       \\
\textit{TrapTube}             (\shortciteauthor{wenkeReasoningGeneralizationRL2019})                                                          & Grid           & Con, D-C             & \textcolor{CBGreen}{D}, \textcolor{CBBlue}{O}, \textcolor{CBPink}{S}                       \\\rowcolor{Gray}
\textit{WordCraft}            (\shortciteauthor{jiang2020wordcraft})                                                                          & LC, Text       & Con, D-C, D-O      & \textcolor{CBOrange}{R}, \textcolor{CBPink}{S}                                              \\
\textit{XLand}                (\shortciteauthor{openendedlearningteamOpenEndedLearningLeads2021})                                             & 3D, LC         & Con, D-C, D-O, PCG & \textcolor{CBGreen}{D}, \textcolor{CBBlue}{O}, \textcolor{CBOrange}{R}, \textcolor{CBPink}{S}   \\\rowcolor{Gray}
\textit{Phy-Q}                (\shortciteauthor{xuePhyQBenchmarkPhysical2021})                                                                & Arcade         & D-C, PCG             & \textcolor{CBPink}{S}                                                                  \\ \bottomrule
\end{tabularx}
\caption{\footnotesize Categorisation of Environments for ZSG. In the \textbf{Style} column, LC stands for Language-Conditioned, ConCon for Continuous Control. In the \textbf{Contexts} column, PCG stands for Procedural Content Generation, Con for continuous, D-C for discrete cardinal and D-O for discrete ordinal. In the \textbf{Variation} column, \textcolor{CBPink}{S}, \textcolor{CBGreen}{D}, \textcolor{CBBlue}{O} and \textcolor{CBOrange}{R} are respectively state, dynamics, observation or reward function variation. In the \textbf{Name} column, \textdagger{} refers to environments that were not originally designed as zero-shot policy transfer benchmarks but could be adapted to be. See main text for a more detailed description of the columns.}\label{table:environments}
\end{table}

\subsection{Environments}\label{subsection:environments}

\paragraph{Categorising Environments That Enable Generalisation}

In \cref{table:environments}, we list the available environments for testing ZSG in RL, as well as summarise each environment's key properties. These environments all provide a non-singleton context set that can be used to create a variety of evaluation protocols. Choosing a specific evaluation protocol then produces a benchmark. We describe the meaning of the columns in \cref{table:environments} here.

\paragraph{Style.} This gives a rough high-level description of the kind of environment.

\paragraph{Contexts.}\label{para:contextset} This describes the context set. In the literature, there are two approaches to designing a context set, and the key difference between these approaches is whether the context-MDP creation is accessible and visible to the researcher. The first, which we refer to as Procedural Content Generation (PCG), relies on a single random seed to determine multiple choices during the context-MDP generation. Here the context set is the set of all supported random seeds. This is a black-box process in which the researcher only chooses a seed.

The second approach provides more direct control over the factors of variation between context-MDPs, and we call these \emph{Controllable} environments. The context set is generally a product of multiple factor spaces, some of which may be discrete (i.e.~a choice between several colour schemes) and some continuous (i.e.\ a friction coefficient in a physical simulation). Borrowing from \shortciteA{keParametricGeneralizationBenchmarking2021}, a distinction between discrete factors of variation is whether they are cardinal (i.e.\ the choices are just a set with no additional structure) or ordinal (i.e.\ the set has additional structure through an ordering). Examples of cardinal factors include different game modes or visual distractions, and ordinal factors are commonly the number of entities of a certain type within the context-MDP\@. All continuous factors are effectively also ordinal factors, as continuity implies an ordering.

Previous literature has defined PCG as any process by which an algorithm produces MDPs given some input \shortcite{risiIncreasingGeneralityMachine2020}, which applies to both kinds of context sets we have described. Throughout the rest of this survey we use ``PCG'' to refer to black-box PCG, which uses a seed as input, and ``controllable'' to refer to environments where the context set directly changes the parameters of interest in the context-MDPs, which could also be seen as ``white-box PCG''. We can understand (black-box) PCG settings as combinations of discrete and continuous factor spaces (i.e.~controllable environments) where the choice of the value in each space is determined by the random generation process. However, only some environments make this more informative parametrisation of the context-MDPs available. In this table, we describe environments where this information is not easily controllable as PCG environments. See \cref{paragraph:procgengen} for a discussion of the downsides of purely PCG approaches.

\paragraph{Variation.} This describes what varies within the set of context MDPs. This could be state-space variation (the initial state distribution and hence implicitly the state space), dynamics variation (the transition function), visual variation (the observation function) or reward function variation. Where the reward varies, the policy often needs to be given some indication of the goal or reward, so that the set of contexts is solvable by a single policy \shortcite{irpanPrincipleUnchangedOptimality2019}.

\subsubsection{Trends In Environments}

There are several trends and patterns shown in \cref{table:environments}, which we draw the reader's attention to here. We describe 55 environments in total and have aimed to be fairly exhaustive.\footnote{As such, if you think there are missing environments, please contact us using the details provided in the author list}

There are a range of different \textbf{Style}s that these environments have, which is beneficial as ZSG methods should themselves be generally applicable across styles if possible. While numerically there is a focus on gridworlds (14, 25\%) and continuous control (13, 24\%) there are well-established benchmarks for arcade styles \shortcite{cobbeLeveragingProceduralGeneration2020} and 3D environments \shortcite{julianiObstacleTowerGeneralization2019}. Looking at \textbf{Context} sets, we see that PCG is heavily used in ZSG environments, featuring in 21 (38\%) environments. Many environments combine PCG components with controllable variation \shortcite{chevalier-boisvertBabyAIPlatformStudy2019,coteTextWorldLearningEnvironment2019,julianiObstacleTowerGeneralization2019,liMetaDriveComposingDiverse2021,openendedlearningteamOpenEndedLearningLeads2021,xuePhyQBenchmarkPhysical2021,fortunatoGeneralizationReinforcementLearners2020,eimerDACBenchBenchmarkLibrary2021,bapstStructuredAgentsPhysical2019}. Most environments have several different kinds of factors of variation within their context set.

There are a lot of differences between environments when looking at the \textbf{Variation} they use. Numerically, state variation is most common (42, 76\%) followed by observation (29, 53\%), and then reward (20, 36\%) and dynamics (19, 35\%). Most environments have multiple different types of variation (34, 62\%), and while there are several environments targeted at just observation variation (10, 18\%) or state variation (9, 16\%), there is only a single environment with solely dynamics variation (RWRL, \shortciteR{dulac-arnoldEmpiricalInvestigationChallenges2021}), and none with solely reward variation. State and Observation variations are often the easiest to engineer, especially with the aid of PCG\@. This is because changing the rendering effects of a simulator, or designing multiple ways the objects in a simulator could be arranged, is generally easier than designing a simulator engine that is parameterisable (for dynamics variation). Creating an environment for reward variation requires further design choices about how to specify the reward function or goal such that the environment satisfies the Principle Of Unchanged Optimality~\shortcite{irpanPrincipleUnchangedOptimality2019}. PCG is often the only good way of generating a large diversity in state variation, and as such is often necessary to create highly varied environments. Only CausalWorld \shortcite{ahmedCausalWorldRoboticManipulation2020} enables easy testing of all forms of variation at once.\footnote{While CARL~\shortcite{benjaminsCARLBenchmarkContextual2021}, MDP Playground~\shortcite{rajanMDPPlaygroundDesign2021}, XLand~\shortcite{openendedlearningteamOpenEndedLearningLeads2021} and TextWorld~\shortcite{coteTextWorldLearningEnvironment2019} are also categorised as containing all forms of variation, CARL is a collection of different environments, none of which have all variation types, XLand is not open-source, and MDP Playground and TextWorld would require significant work to construct a meaningful evaluation protocol which varies along all factors. Further, MDP Playground does not provide a method for describing the changed reward function to the agent, and there is uncertainty in how to interpret observation variation in text-based games like TextWorld.}

There are several clusters that can be pointed out in the collection of benchmarks: There are several PCG state-varying gridworld environments (MiniGrid, BabyAI, Crafter, Rogue-gym, MarsExplorer, NLE, MiniHack;  \shortciteR{chevalier-boisvertMinimalisticGridworldEnvironment2021,chevalier-boisvertBabyAIPlatformStudy2019,hafnerBenchmarkingSpectrumAgent2021,kanagawaRogueGymNewChallenge2019,koutrasMarsExplorerExplorationUnknown2021,kuttlerNetHackLearningEnvironment2020,samvelyanMiniHackPlanetSandbox2021}), non-PCG observation-varying continuous control environments (RoboSuite, DMC-Remastered, DMC-GB, DCS, KitchenShift, NaturalEnvs MuJoCo;  \shortciteR{fanSECANTSelfExpertCloning2021,grigsbyMeasuringVisualGeneralization2020,hansenGeneralizationReinforcementLearning2021,stoneDistractingControlSuite2021,xingEvaluatingGeneralizationPolicy2021,zhangNaturalEnvironmentBenchmarks2018}), and multi-task continuous control benchmarks which could be adapted to ZSG (CausalWorld, RLBench, Meta-world;  \shortciteR{ahmedCausalWorldRoboticManipulation2020,jamesRLBenchRobotLearning2019,yuMetaworldBenchmarkEvaluation2019}).

\subsection{Evaluation Protocols For Zsg}\label{subsection:evalprotocol}
As discussed, a benchmark is the combination of an environment and an evaluation protocol. Each environment supports a range of evaluating protocols determined by the context set, and often there are protocols recommended by the environment creators. In this section, we discuss the protocols and the differences between them. An evaluation protocol specifies the training and testing context sets, any restrictions on sampling from the training set during training, and the number of samples allowed from the training environment.

An important first attribute that varies between evaluation protocols is \emph{context-efficiency}. This is analogous to sample efficiency, where only a certain number of samples are allowed during training, but instead, we place restrictions on the number of \emph{contexts}. This ranges from a single context to a small number of contexts, to the entire context set.

\begin{figure}[t!]
\centering
\includegraphics[width = \textwidth]{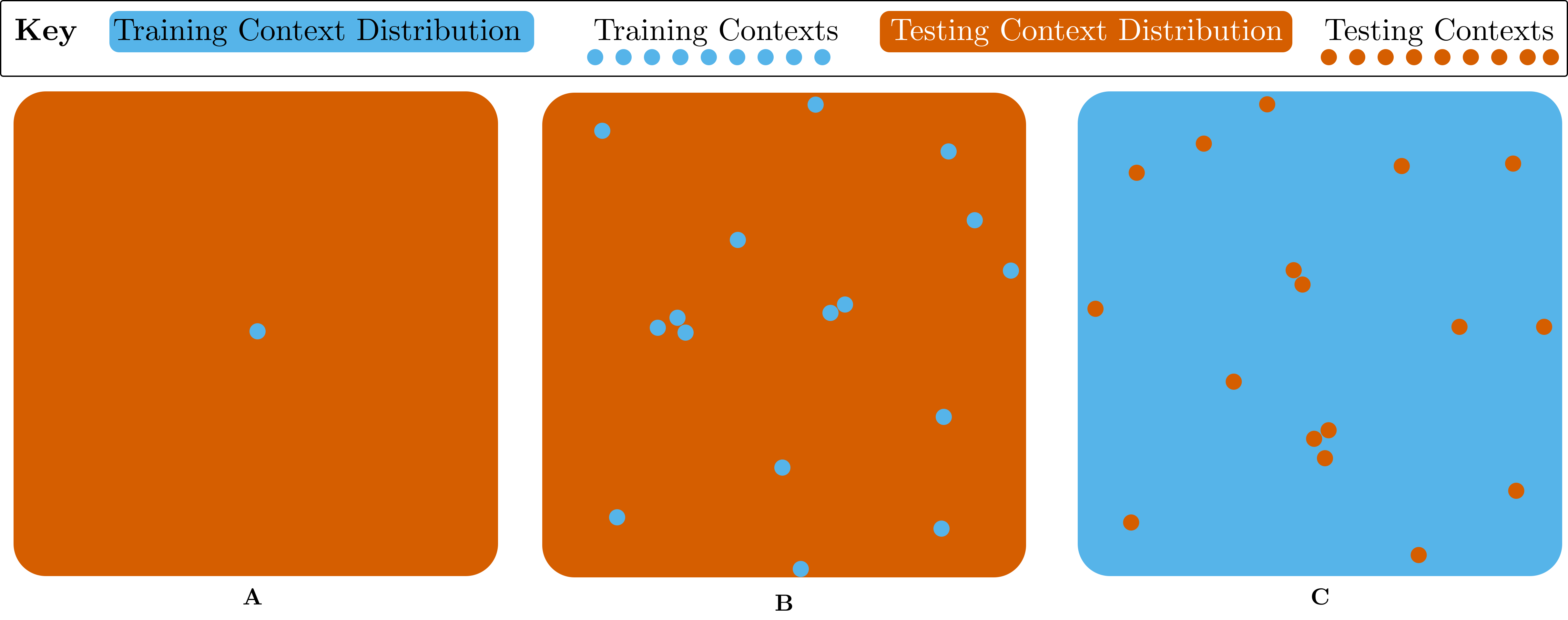}
\caption{\textbf{Visualisation of Evaluation Protocols for PCG Environments}. \textbf{A} is a single training context, and the whole context set for testing. \textbf{B} uses a small collection of training contexts randomly sampled from the context set and the entire space for testing. \textbf{C} effectively reverses this, using the entire context set for training apart from several randomly sampled held-out contexts that are used for testing. The lack of axes indicates that these sets have no structure.}\label{fig:evalprotpcg}
\end{figure}

\paragraph{PCG Evaluation Protocols.} In fact, in purely PCG environments, the only meaningful factor of variation between evaluation protocols is the context efficiency restriction. As we have no control over the factors of variation apart from sampling random seeds, the only choice we have is how many contexts to use for training. Further, the only meaningful testing context set is the full distribution, as taking a random sample from it (the only other option) would just be an approximation of the performance on the full distribution. This limitation of PCG environments is discussed further below (\cref{paragraph:procgengen}).

This gives three classes of evaluation protocol for PCG environments, as determined by their training context set: A single context, a small set of contexts, or the full context set. These are visualised in \cref{fig:evalprotpcg} \textbf{A}, \textbf{B} and \textbf{C} respectively. There are not any examples of protocol \textbf{A} (for purely PCG environments), likely due to the high difficulty of such a challenge. For protocol \textbf{B}, while ``a small set of contexts'' is imprecise, the relevant point is that this set is meaningfully different from the full context set: it is possible to overfit to this set without getting good performance on the testing set. Examples of this protocol include both modes of OpenAI Procgen~\shortcite{cobbeLeveragingProceduralGeneration2020}, RogueGym~\shortcite{kanagawaRogueGymNewChallenge2019}, some uses of JumpingFromPixels~\shortcite{tachetLearningInvariancesPolicy2020} and MarsExplorer~\shortcite{koutrasMarsExplorerExplorationUnknown2021}.

Protocol \textbf{C} is commonly used among PCG environments that are not explicitly targeted at ZSG (MiniGrid, NLE, MiniHack, Alchemy;  \shortciteR{chevalier-boisvertMinimalisticGridworldEnvironment2021,kuttlerNetHackLearningEnvironment2020,samvelyanMiniHackPlanetSandbox2021,wangAlchemyStructuredTask2021}). The testing context set consists of seeds held out from the training set, and otherwise during training the full context set is used. This protocol effectively tests for more robust RL optimisation improvements but does not test for zero-shot generalisation beyond avoiding memorising. While this protocol only tests for ZSG in a weak sense, it still matches a wider variety of real-world deployment scenarios than the previous standard in RL, where the evaluation of the policy is performed on the environment it is trained in, and so we believe it should be the standard evaluation protocol in RL (not just in ZSG), and the previous standard should be considered a special case.

\paragraph{Controllable Environment Evaluation Protocols.} Many environments do not use only PCG and have factors of variation that can be controlled by the user of the environment. In these controllable environments, there is a much wider range of possible evaluation protocols.

\begin{figure}[b!]
\centering{}
\includegraphics[width = \textwidth]{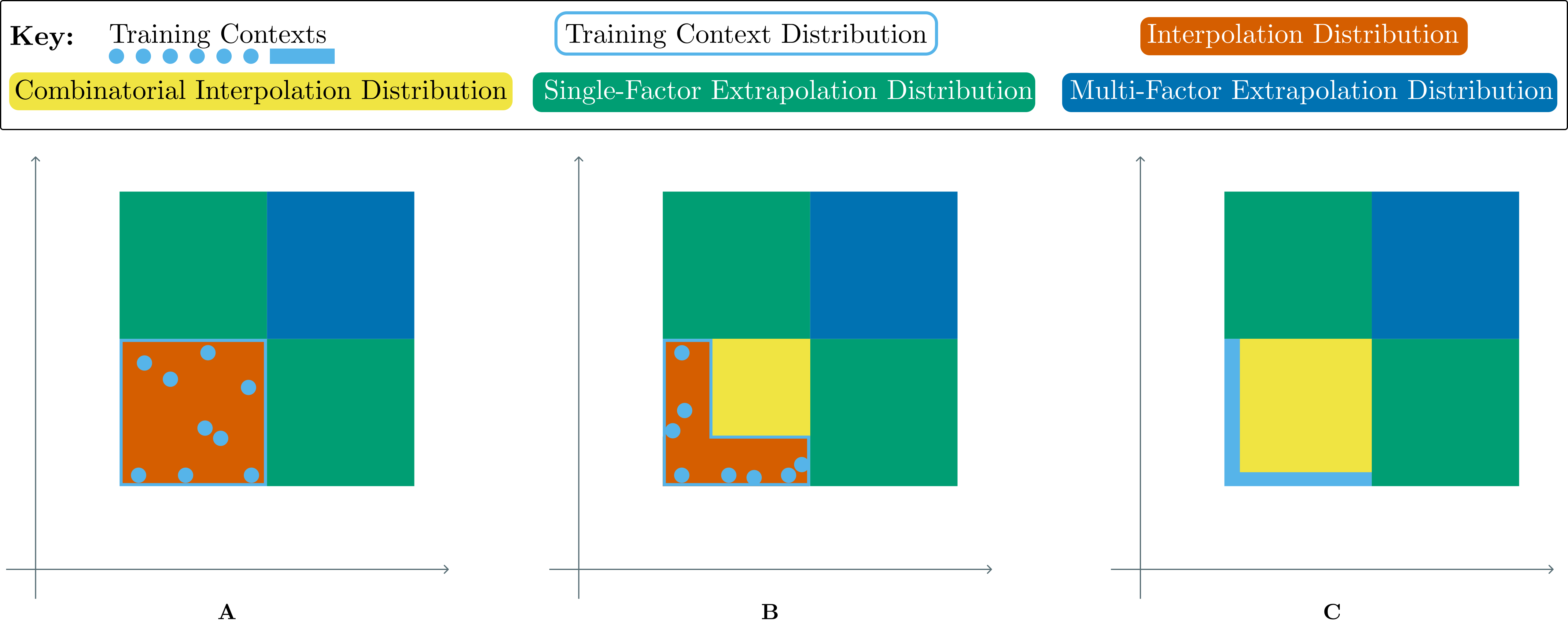}
\caption{\textbf{Visualisation of Evaluation Protocols for Controllable Environments}. Each diagram visualises one possible training context set (blue), and multiple possible testing context sets (all other colours). In \textbf{A} we choose the range for each factor of variation independently for the training distribution, resulting in a convex shape for this distribution. In this setting, possible testing distributions can either be interpolation (red), extrapolation along a single factor (either green square) or extrapolation along both factors (blue). In \textbf{B} and \textbf{C} the ranges for each factor of variation are linked together, resulting in a non-convex shape for the training distribution. This allows an additional type of generalisation to be tested, combinatorial interpolation (yellow), where the factors take values seen during training independently, but in unseen combinations. We continue to have the previous interpolation and extrapolation testing distributions. The difference from \textbf{B} to \textbf{C} is in the width of the training distribution in the axes along which we expect the agent to generalise. In \textbf{C} the policy will not be able to learn that the two factors can vary independently at all, making all forms of generalisation harder. Note that in actual environments and real-world settings it is likely this space will be higher than two dimensions and contain non-continuous and non-ordinal axes. The axes indicate that in this setting we have control over these factors of variation, in contrast to \protect{\cref{fig:evalprotpcg}}.}\label{fig:evalprotcontrol}
\end{figure}

The choice in PCG protocols -- between a single context, a small set, or the entire range of contexts -- transfers to the choice for each factor of variation in a controllable environment. For each factor, we can choose one of these options for the training context set, and then choose to sample either within or outside this range for the testing context set. The range of options is visualised in \cref{fig:evalprotcontrol}.

Making this choice for each factor independently gives us a convex training context set within the full context set (\cref{fig:evalprotcontrol} \textbf{A}). For testing, each factor can then be either inside or outside this convex set (often respectively referred to as interpolation and extrapolation). The number of factors chosen to be extrapolating contributes to the difficulty of the evaluation protocol.

However, if we create correlations or links between values of factors during training, we can get a non-convex training context set within the full context set (\cref{fig:evalprotcontrol} \textbf{B}). Each possible testing context can either be within the training context set (fully interpolation), within the set formed by taking the convex hull of the non-convex training context set (combinatorial interpolation), or fully outside the convex hull (extrapolation). Combinatorial interpolation tests the ability of an agent to exhibit \emph{systematicity}, a form of compositional ZSG discussed in \cref{subsection:backgroundgen}. For ordinal factors, we can also choose disjoint ranges, which allows us to test interpolation along a single axis (i.e.~taking values between the two ranges). Note that when discussing convex hulls, this only applies to factors of variation that are continuous or discrete-ordinal; for cardinal factors of variation, the convex hull just includes those values sampled during training.

For example, consider a policy trained in a CMDP where the context set consists of values for friction and gravity strength. During training, the environments instances have either a friction coefficient between 0.5 and 1 but gravity fixed at 1, or gravity strength ranging between 0.5 and 1 but friction fixed at 1 (the light blue line in \cref{fig:evalprotcontrol} \textbf{C}). Testing contexts which take friction and gravity values within the training distribution are full interpolation (e.g.~$(f=0.5, g=1), (f=1, g=1)$), contexts which take values for friction and gravity which have been seen independently but not in combination are combinatorial interpolation (e.g.\ $(f=0.5, g=0.5), (f=0.5, g=0.9)$, the yellow area), and contexts which take values for friction and gravity which are outside the seen ranges during training are full extrapolation (e.g.\ $(f=0.2, g=0.5), (f=1.1, g=1.5)$, either the dark blue or green areas). 

We can still consider the number of contexts within the training context set, which controls the density of the training context set, given its shape. When testing for extrapolation we can also vary the ``width'' of the training context set on the axis of variation along which extrapolation is being tested (\cref{fig:evalprotcontrol} \textbf{B} vs \textbf{C}). These tests evaluate the agent's ability to exhibit \emph{productivity} (\cref{subsection:backgroundgen}).  Of course, ZSG will be easier if there is a wide diversity of values for this factor at training time, even if the values at test time are still outside this set. For example, if we are testing whether a policy can generalise to novel amounts of previously seen objects, then we should expect the policy to perform better if it has seen different amounts during training, as opposed to only having seen a single amount of the object during training. To expand on the friction and gravity example, during training the policy never sees gravity and friction varying together, which makes it much more difficult to generalise to the testing contexts. If gravity and friction did vary together during training then this would make generalisation easier.

A notable point in this space is that of a single training context and a wide variety of testing contexts. This protocol tests for a strong form of ZSG, where the policy must be able to extrapolate to unseen contexts at test time. Because of the difficulty of this problem, benchmarks with this evaluation protocol focus on visual variation: the policy needs to be robust to different observation functions on the same underlying MDP \shortcite{grigsbyMeasuringVisualGeneralization2020,stoneDistractingControlSuite2021,xingEvaluatingGeneralizationPolicy2021}. The protocol is often motivated by the sim-to-real problem, where we expect an agent trained in a single simulation to be robust to multiple visually different real-world settings at deployment time.

Beyond this single point, it is challenging to draw any more meaningful categorisation from the current array of evaluation protocols. Generally, each one is motivated by a specific problem setting or characteristic of human reasoning which we believe RL agents should be able to solve or have respectively.

\subsection{Discussion}\label{subsection:benchmarkdiscuss}

There are several comments, insights and conclusions that can be gained from surveying the breadth of ZSG benchmarks, which we raise here. 

\paragraph{Non-visual Generalisation.} If testing for non-visual types of generalisation, then visually simple domains such as MiniHack \shortcite{samvelyanMiniHackPlanetSandbox2021} and NLE \shortcite{kuttlerNetHackLearningEnvironment2020} should be used. These environments contain enough complexity to test for many types and strengths of non-visual generalisation but save on computation due to the lack of complex visual processing required. There are many real-world problem settings where no visual processing is required, such as systems control and recommender systems. Representation learning is still a problem in these non-visual domains, as many of them have a large variety of entities and scenarios which representations and hence policies need to generalise across. 

\paragraph{DeepMind Control Suite Variants.} A sub-category of ZSG benchmarks unto itself is the selection of DeepMind Control Suite~\shortcite{tassaDeepMindControlSuite2018} variants: DMC-Remastered, DMC-Generalisation Benchmark, Distracting Control Suite, Natural Environments \shortcite{grigsbyMeasuringVisualGeneralization2020,hansenGeneralizationReinforcementLearning2021,stoneDistractingControlSuite2021,zhangNaturalEnvironmentBenchmarks2018}. All these environments focus on visual generalisation and sample efficiency, require learning continuous control policies from pixels and introduce visual distractors that the policy should be invariant to which are either available during training or only present at deployment. We believe that Distracting Control Suite \shortcite{stoneDistractingControlSuite2021} is the most fully-featured variant in this space, as it features the broadest set of variations, the hardest combinations of which are unsolvable by current methods.

\paragraph{Unintentional Generalisation Benchmarks.} Some environments listed in \cref{table:environments} were not originally intended as ZSG benchmarks. For example,~\shortciteA{toschToyboxSuiteEnvironments2019} presents three highly parameterisable versions of Atari games and uses them to perform post hoc analysis of agents trained on a single variant. Some environments are not targeted at zero-shot policy transfer (CausalWorld, RWRL, RLBench, Alchemy, Meta-world \shortcite{ahmedCausalWorldRoboticManipulation2020,dulac-arnoldEmpiricalInvestigationChallenges2021,jamesRLBenchRobotLearning2019,wangAlchemyStructuredTask2021,yuMetaworldBenchmarkEvaluation2019}), but could be adapted to such a scenario with a different evaluation protocol. More generally, all environments provide a context set, and many then propose specific evaluation protocols, but other protocols could be used as long as they were well-justified. This flexibility has downsides, as different methods can be evaluated on subtly different evaluation protocols which may favour one over another. We recommend being explicit when using these benchmarks in exactly which protocol is being used and comparing with evaluations of previous methods. Using a standard protocol aids reproducibility.

\paragraph{The Downsides of Procedural Content Generation for Zero-shot Generalisation.}\label{paragraph:procgengen}
Many environments make use of procedural content generation (PCG) for creating a variety of context-MDPs. In these environments, the context set is the set of random seeds used for the PCG and has no additional structure with which to control the variation between context-MDPs.

This means that while PCG is a useful tool for creating a large set of context-MDPs, there is a downside to purely PCG-based environments: the range of evaluation protocols supported by these environments is limited to different sizes of the training context set. Measuring zero-shot generalisation along specific factors of variation is impossible without significant effort either labelling generated levels or unravelling the PCG to expose the underlying parametrisation which captures these factors. Often more effort is required to enable setting these factors to specific values, as opposed to just revealing their values for generated levels. Hence, PCG benchmarks are testing for a ``general'' form of ZSG and RL optimisation, but do not enable more targeted evaluation of specific types of ZSG\@. This means making research progress on specific problems is difficult, as focusing on the specific bottleneck in isolation is hard.

An interesting compromise, which is struck by several environments, is to have some low-level portion of the environment procedurally generated, but still have many factors of variation under the control of the researcher. For example, Obstacle Tower \shortcite{julianiObstacleTowerGeneralization2019} has procedurally generated level layouts, but the visual features (and to some extent the layout complexity) can be controlled. Another example is MiniHack \shortcite{samvelyanMiniHackPlanetSandbox2021}, where entire MDPs can be specified from scratch in a rich description language, and PCG can fill in any components if required. These both enable more targeted types of experimentation. We believe this kind of combined PCG and controllable environment is the best approach for designing future environments; some usage of PCG will be necessary to generate sufficient variety in the environments (especially the state space), and if the control is fine-grained enough to enable precise scientific experimentation, then the environment will still be useful for disentangling progress in ZSG\@.

\paragraph{Compositional Generalisation in Contextual MDPs.}\label{paragraph:compositionalstructure}
Compositional generalisation is a key point of interest for many researchers (see \cref{subsection:backgroundgen}). In \emph{controllable} environments, different evaluation protocols enable us to test for some of the forms of compositional generalisation introduced in \cref{subsection:backgroundgen} \shortcite{hupkesCompositionalityDecomposedHow2020}:
(1) Systematicity can be evaluated using a multidimensional context set, and testing on novel combinations of the context dimensions not seen at training time (combinatorial interpolation in \cref{fig:evalprotcontrol}). (2) Productivity can be evaluated with ordinal or continuous factors, measuring the ability to perform well in environment instances with context values beyond those seen at training time (either type of extrapolation in \cref{fig:evalprotcontrol}). If dealing with CMDPs where the context space is partially language, as in language-conditioned RL \shortcite{luketinaSurveyReinforcementLearning2019}, then the evaluations discussed by \shortciteA{hupkesCompositionalityDecomposedHow2020} can be directly applied to the language space. Crucially, a controllable environment with a structured context space is required to test these forms of compositional generalisation, and to ensure that the agent is seeing truly novel combinations at test time; this is difficult to validate in PCG environments like OpenAI Procgen \shortcite{cobbeLeveragingProceduralGeneration2020} or NLE \shortcite{kuttlerNetHackLearningEnvironment2020}.

The other forms of compositional generalisation in \shortcite{hupkesCompositionalityDecomposedHow2020} require additional structure not captured by the choices of evaluation protocol, and we describe what testing these forms could entail here: (3) substitutivity through the use of synonyms (in language) or equivalent objects and tools; (4) locality through comparing the interpretation of an agent given command A and command B separately vs.\ the combination of A + B and if those interpretations are different; and (5) overgeneralisation through how the agent responds to exceptions in language or rules of the environment.

\paragraph{What Generalisation Can We Expect?}\label{paragraph:genexpect}

In \cref{subsection:evalprotocol} we discussed a variety of different possible evaluation protocols for different styles of ZSG benchmark. However, it is a different question in which protocols we can expect reasonable performance. For example, it is unreasonable to expect a standard RL policy trained \emph{tabula rasa} on a single level of NLE \shortcite{kuttlerNetHackLearningEnvironment2020} to generalise to dissimilar levels, as it might encounter entirely unseen entities or very out-of-distribution combinations of entities and level layouts. Each evaluation protocol measures a different kind of ZSG strength, and they hence form a kind of partial ordering where ``easier'' evaluation protocols come before ``harder'' protocols. We outline this ordering here:
\begin{itemize}
	\item Increasing the number of samples can make an evaluation protocol easier, but often only to a point: more samples are unlikely to bring greater variety which is needed for zero-shot generalisation. Increasing the number of contexts (while keeping the shape of the context set the same) also makes an evaluation protocol easier. Even a small amount of additional variety can improve performance.
	\item The number of factors of variation which are extrapolating or combinatorially interpolating in the testing context set can also be varied. The more there are, the more difficult the evaluation protocol. Further, the width of the range of values that the extrapolating factors take at training time can vary. This is linked to the number of contexts but is also related to the variety available during training time along these axes of variation.
	\item Following \shortciteA{keParametricGeneralizationBenchmarking2021}, we consider the difficulty of interpolating and extrapolating along different types of factors of variation. Interpolation along ordinal axes is likely the easiest, followed by cardinal axes interpolation (which happens through unseen combinations of seen values for a cardinal axis combined with any other axis), and then extrapolation along ordinal axes. Finally, extrapolation along cardinal axes is the most difficult.
\end{itemize}

As the difficulty of an evaluation protocol increases, it becomes less likely that standard RL approaches will get good performance. In more difficult protocols which involve extrapolation of some form, zero-shot generalisation is unlikely to occur at all with standard RL methods, as there is no reason to expect a policy to generalise correctly to entirely unseen values. That does not mean that this type of generalisation is impossible: it just makes clear the fact that to achieve it, more than standard RL methods will be needed. That is, methods incorporating prior knowledge\footnote{\url{http://betr-rl.ml/2020/}} such as transfer from related environments \shortcite{zhuTransferLearningDeep2021}; strong inductive biases~\shortcite{zambaldiDeepReinforcementLearning2018} or assumptions about the variation; or utilising online adaptation~\shortcite{hansenSelfSupervisedPolicyAdaptation2021,zintgrafVariBADVeryGood2020} will be necessary to produce policies that generalise in this way.

\section{Methods For Zero-shot Generalisation In Reinforcement Learning}\label{section:methods}

We now classify methods that tackle ZSG in RL\@. The problem of ZSG occurs when the training and the testing context sets are not identical. There are many types of ZSG problems (as described in more detail in \cref{section:benchmarks}), and hence there are many different styles of methods. We categorise the methods into those that try and increase the similarity between training and testing data and objective (\cref{subsection:similarity}), those that explicitly aim to handle differences between training and testing environments (\cref{subsection:difference}), and those that target RL-specific issues or optimisation improvements which aid ZSG performance (\cref{subsection:rlspecific}).

\begin{figure}[t]
\centering
\includegraphics[width = \textwidth]{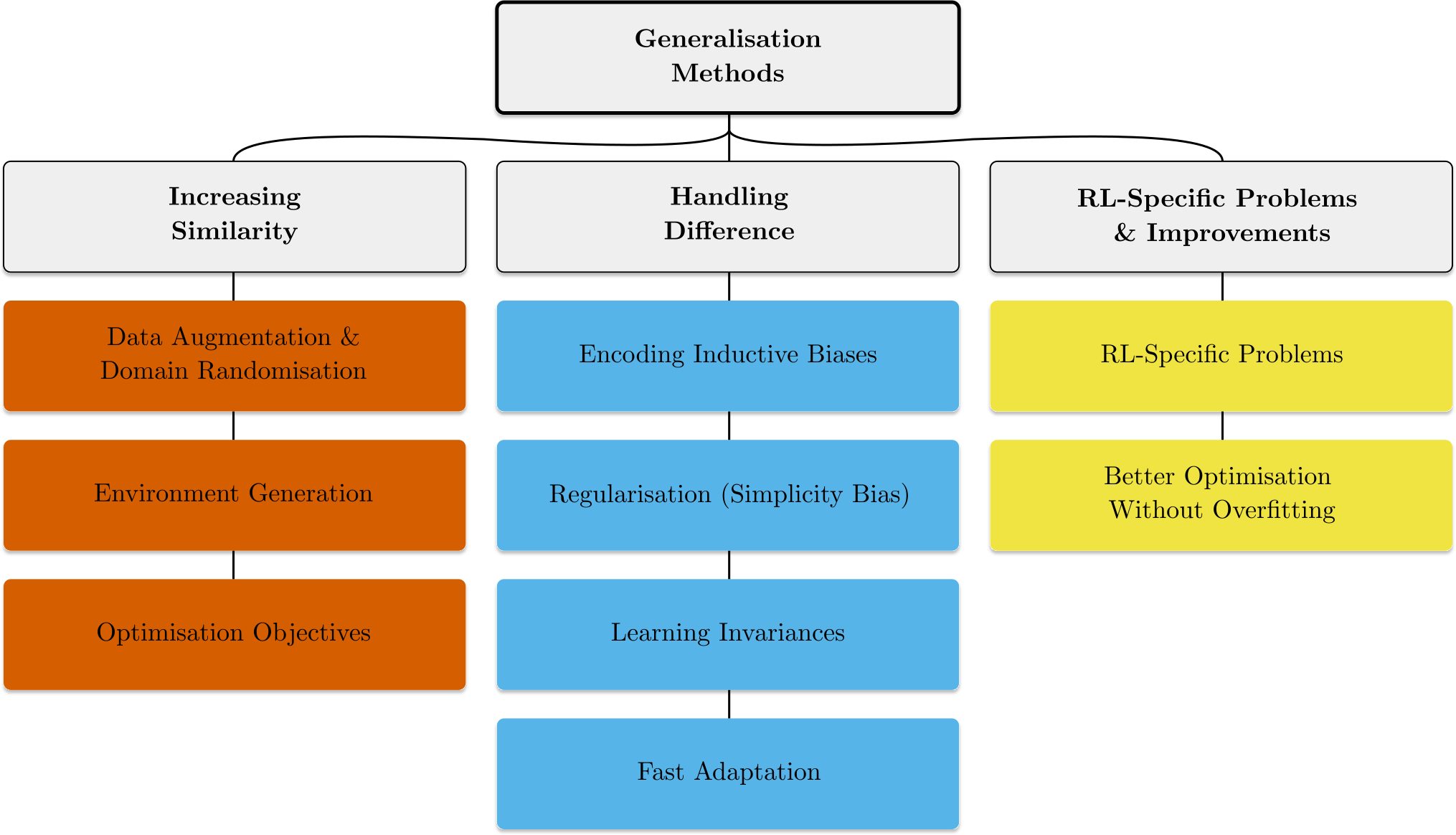}
\caption{\textbf{Categorisation of methods for tackling zero-shot generalisation in reinforcement learning}}\label{fig:methods}
\end{figure}

See \cref{fig:methods} for a diagram of this categorisation, and \cref{table:methodsone,table:methodstwo} for a table classifying methods by their approach, the environment variation they were evaluated on, and whether they mostly change the environment, loss function or architecture. Performing this comprehensive classification enables us to see the under-explored areas within ZSG research, and we discuss future work for methods in \cref{subsection:futuremethods}.

\subsection{Increasing Similarity Between Training And Testing}\label{subsection:similarity}

All else being equal, the more similar the training and testing environments are, the smaller the generalisation gap and the higher the test time performance. This similarity can be increased by designing the training environment to be as close to the testing environment as possible. Assuming this has been done, in this section, we cover methods that make the data and objective being used to learn the policy during training closer to that which would be used if we were optimising on the testing environment.

\subsubsection{Data Augmentation and Domain Randomisation}
Two natural ways to make training and testing data more similar are data augmentation \shortcite{shortenSurveyImageData2019} and domain randomisation \shortcite{tobinDomainRandomizationTransferring2017,sadeghiCAD2RLRealSingleImage2017,pengSimtoRealTransferRobotic2018}. This is especially effective when the variation between the training and testing environments is known, as then a data augmentation or domain randomisation can be used which captures this variation. In practice there is only so far this approach can go, as stronger types of variation often cannot be captured by this simple method.

Data augmentation (DA) can be viewed in two ways. First, the augmented data points are seen as additional data to train the model. This interpretation is what causes us to classify DA techniques as trying to increase the similarity between training and testing data. In the second view, DA can be used to enforce the learning of an invariance, by regularising the model to have the same output (or the same internal representations) for different augmented data points. In this view, DA is more with encoding inductive biases, which we cover in \cref{subsubsection:inductivebias}. We include all DA work in this section for clarity.

There are many examples of using DA in RL, although not all of them are targeted at ZSG performance. \shortciteA[UCB-DrAC]{raileanuAutomaticDataAugmentation2021} adapts the DA technique DrQ \shortcite{kostrikovImageAugmentationAll2021} to an actor-critic setting \shortcite[PPO]{schulmanProximalPolicyOptimization2017} and introduces a method for automatically picking the best augmentation during training. \shortciteA[Mixreg]{wangImprovingGeneralizationReinforcement2020} adapts mixup \shortcite{zhangMixupEmpiricalRisk2018} to the RL setting, which encourages the policy to be linear in its outputs with respect to mixtures of possible inputs.~\shortciteA[PAADA]{zhangGeneralizationReinforcementLearning2021} use adversarial DA combined with mixup. \shortciteA[RandFM]{leeNetworkRandomizationSimple2020} uses a randomised convolutional layer at the start of the network to improve robustness to a wide variety of visual inputs. \shortciteA[MixStyle]{zhouDomainGeneralizationMixStyle2021} mixes style statistics across spatial dimensions in CNNs for increased data diversity. All these methods \shortcite{raileanuAutomaticDataAugmentation2021,wangImprovingGeneralizationReinforcement2020,zhangGeneralizationReinforcementLearning2021,leeNetworkRandomizationSimple2020,zhouDomainGeneralizationMixStyle2021} show improved performance on CoinRun \shortcite{cobbeQuantifyingGeneralizationReinforcement2019} or OpenAI Procgen \shortcite{cobbeLeveragingProceduralGeneration2020} by improving both training and testing performance, and some also show gains on other benchmarks such as visually distracting DeepMind Control (DMC) variants. \shortciteA[SODA]{hansenGeneralizationReinforcementLearning2021} uses similar augmentations as before but only to learn a more robust image encoder, while the policy is trained on non-augmented data, demonstrating good performance on DMC-GB \shortcite{hansenGeneralizationReinforcementLearning2021}. \shortciteA[RCAN]{jamesSimtoRealSimtoSimDataefficient2019} use DA to learn a visual mapping from any different observation back to a canonical observation of the same state, and then train a policy on this canonical observation. They show improved sim-to-real performance on a robotic grasping task.

\shortciteA[InDA,ExDA]{koTimeMattersUsing2021} show that the time at which the augmentations are applied is important for the performance: some augmentations help during training, whereas others only need to be applied to regularise the final policy. \shortciteA[SECANT]{fanSECANTSelfExpertCloning2021} introduce a method for combining DA with policy distillation. As training on strong augmentations can hinder performance, they first train on weak augmentations to get an expert policy, which they then distil into a student policy trained with strong augmentations. \shortciteA[SVEA]{hansenStabilizingDeepQLearning2021} argues that DA when naively applied increases the variance of Q-value targets, making learning less stable and efficient. They introduce adjustments to the standard data-augmentation protocol by only applying augmentations to specific components at specific points during the calculation of the loss function, evaluating performance on DMC-GB \shortcite{hansenGeneralizationReinforcementLearning2021}. These works show that in RL settings the choice of when to apply augmentations and what type of augmentations to apply is non-trivial, as the performance of the model during training impacts the final performance by changing the data the policy learns from.

Domain Randomisation (DR) is the practice of randomising the environment across a distribution of parametrisations, aiming for the testing environment to be covered by the distribution of environments trained on. Fundamentally DR is just the creation of a non-singleton training context set, and then randomly sampling from this set. \shortciteA{tobinDomainRandomizationTransferring2017},~\shortciteA{sadeghiCAD2RLRealSingleImage2017} and~\shortciteA{pengSimtoRealTransferRobotic2018} introduced this idea in the setting of sim-to-real transfer in robotics. Much work has been done on different types of DR, so we cover just a sample here.~\shortciteA{openaiSolvingRubikCube2019} described Automatic Domain Randomisation: instead of sampling possible environment parametrisations uniformly at random, this method dynamically adjusts the distribution in response to the agent's current performance.~\shortciteA[Minimax DSAC]{renImprovingGeneralizationReinforcement2020} use adversarial training to learn the DR for improved robustness.~\shortciteA[P2PDRL]{zhaoRobustDomainRandomised2020} improve DR though peer-to-peer distillation. \shortciteA[DDL]{wellmerDropoutDreamLand2021} learns a world model in which to train a policy and then applies dropout to the recurrent network within the world model, effectively performing DR in imagination. Finally, as procedural content generation \shortcite{risiIncreasingGeneralityMachine2020} is a method for generating a non-zero context set, it can be seen as a form of DR\@. Works on DR generally leverage the possibility of using a non-uniform context distribution, which possibly varies during training.

In both DA and DR approaches, as the training environment instances are increasingly augmented or randomised, optimisation becomes increasingly difficult, which often makes these methods much less sample-efficient. This is the motivation behind \shortciteS{jamesSimtoRealSimtoSimDataefficient2019,hansenGeneralizationReinforcementLearning2021,fanSECANTSelfExpertCloning2021} work and other works, all of which train an RL policy on a non-randomised or only weakly randomised environment while using other techniques such as supervised or self-supervised learning to train a robust visual encoder.

Finally, note that most of the DA techniques are focused on visual variation in the context set, as that is the easiest variation to produce useful augmentations for. Some DR work focuses on dynamics variation as a way of tackling the sim-to-real problem, where it is assumed that the dynamics will change between training (simulation) and testing (reality).

\subsubsection{Environment Generation}\label{subsubsection:environmentgen}

While DR and PCG produce context-MDPs within a pre-determined context set, it is normally assumed that all the context-MDPs are solvable. However, in some settings, it is unknown how to sample from the set of all \emph{solvable} context-MDPs. For example, consider a simple gridworld maze environment, where the context set consists of all possible block layouts on the grid; some configuration of block placements will result in unsolvable mazes. Further, many block configurations are not useful for training: they may produce trivially easy mazes. To solve these problems we can \emph{learn} to generate new levels (sample new contexts) on which to train the agent, such that we can be sure these context-MDPs are solvable and useful training instances. We want a distribution over context-MDPs which is closer to the testing context set, which likely has only solvable context-MDPs. This is known as environment generation.

\shortciteA{wangPairedOpenEndedTrailblazer2019} introduced Paired Open-Ended Trailblazer (POET), a method for jointly evolving context-MDPs and policies which solve those MDPs, aiming for a policy that can solve a wide variety of context-MDPs. They produce policies that solve unseen level instances reliably and perform better than training from scratch or a naive curriculum. \shortciteA{wangEnhancedPOETOpenEnded2020} built on POET, introducing improvements to the open-ended algorithm including a measure of how novel generated context-MDPs are and a generic measure of how much a system exhibits open-ended innovation. These additions improve the diversity and complexity of context-MDPs generated. 

\shortciteA{dennisEmergentComplexityZeroshot2021} introduces the framework of Unsupervised Environment Design (UED), similar to POET, in which the task is to generate context-MDPs in an unsupervised manner, which are then used to train a policy. The aim is to improve ZSG to unseen tasks either within or outside the environment's context-MDP space. Their method PAIRED outperforms standard DR and a method analogous to POET in the grid-world setting described above, as measured by the zero-shot generalisation performance to unseen levels. \shortciteA{jiangReplayGuidedAdversarialEnvironment2021} extended the formal framework of UED, combining it with Prioritized Level Replay \shortcite[PLR]{jiangPrioritizedLevelReplay2021}, and motivates understanding PLR as an environment generation algorithm. This combined method shows improved performance in terms of zero-shot generalisation to unseen out-of-distribution tasks in both gridworld mazes and 2D car-racing tracks. We summarise PLR in the next section.

Environment generation and DR are both methods for adjusting the context distribution over some context set provided by the environment. Environment generation tends to learn this sampling procedure, and is targeted at environments where the context set is unstructured such that not all context-MDPs are solvable or useful for training, whereas DR work often uses hard-coded heuristics or non-parametric learning approaches to adjust the context distribution, and focuses on settings where the domains are all solvable but possibly have different difficulties or learning potentials. Both can often also be seen as a form of automatic curriculum learning \shortcite{portelasAutomaticCurriculumLearning2020}, especially if the context distribution is changing during training and adapting to the agent's performance.

This area is very new and we expect there to be more research soon. However, it does require access to an environment where context-MDPs can be generated at a fairly fine level of detail. Environment generation methods can target ZSG over any kind of variation, as long as that kind of variation is present in the output space of the context-MDP generator. Current methods focus on state-space variation, as that is the most intuitive to formulate as a context set within which the generator can produce specific MDPs.

\subsubsection{Optimisation Objectives}

It is sometimes possible to change our optimisation objective (explicitly or implicitly) to one which better aligns with testing performance. 

Changing the distribution over which the training objective is calculated can be seen as implicitly changing the optimisation objective. An initial example in this area applied to improving ZSG is PLR \shortcite{jiangPrioritizedLevelReplay2021}, in which the sampling distribution over levels is changed to increase the learning efficiency and zero-shot generalisation of the trained policy. They show both increased training and testing performance on OpenAI Procgen, and the method effectively forms a rough curriculum over levels, enabling more sample-efficient learning while also ensuring that no context-MDP's performance is too low. 

Methods for Robust RL (RRL) are also targeted at the ZSG problem, and work by changing the optimisation objective of the RL problem. These methods take a worst-case optimisation approach, maximising the minimum performance over a set of possible environment perturbations (which can be understood as different context-MDPs), and are focused on improving ZSG to unseen dynamics. \shortciteA{chenOverviewRobustReinforcement2020} gave an overview and introduction to the field. \shortciteA[WR$^2$L]{abdullahWassersteinRobustReinforcement2019} optimised the worst-case performance using a Wasserstein ball around the transition function to define the perturbation set. \shortciteA[SRE-MPO]{mankowitzRobustReinforcementLearning2020} incorporates RLL into MPO \shortcite{abdolmalekiMaximumPosterioriPolicy2018} and shows improved performance on the RWRL benchmark \shortcite{dulac-arnoldEmpiricalInvestigationChallenges2021}. \shortciteA[RARL]{pintoRobustAdversarialReinforcement2017} also (implicitly) optimises a robust RL objective through the use of an adversary which is trained to pick the worst perturbations to the transition function. This can also be seen as an adversarial domain randomisation technique.

\subsection{Handling Differences Between Training And Testing}\label{subsection:difference}

One way of conceptualising why policies do not transfer perfectly at test time is due to the differences between the two environments: the trained model will learn to rely on features during training that change in the testing environment and performance then suffers. In this section, we review methods that try and explicitly handle the possible differences between the features of the training and testing environments.

\subsubsection{Encoding Inductive Biases}\label{subsubsection:inductivebias}

If we know how features change between the training and testing context-MDPs, we can use inductive biases to encourage or ensure the model does not rely on features that we expect to change: The policy should only rely on features which will behave similarly in both the training and testing environments. For example, if we know colour varies between training and testing, and colour is irrelevant to the task, then we can remove colour from the visual input before processing. Simple changes like this tend not to be worthy of separate papers, but they are still important to consider in real-world problem scenarios. 

IDAAC \shortcite{raileanuDecouplingValuePolicy2021} adds an adversarial regularisation term which encourages the internal representations of the policy not to be predictive of time within an episode. This invariance is useful for OpenAI Procgen~\shortcite{cobbeLeveragingProceduralGeneration2020}, as timestep is irrelevant for the optimal policy but could be used to overfit to the training set of levels.~\shortciteA[DARLA]{higginsDARLAImprovingZeroShot2018} uses $\beta$-VAEs~\shortcite{higginsBetaVAELearningBasic2016} to encode the inductive bias of disentanglement into the representations of the policy, improving zero-shot performance on various visual variations.~\shortciteA[NAP]{vlastelicaNeuroalgorithmicPoliciesEnable2021} incorporates a black-box shortest-path solver to improve ZSG performance in hard navigation problems.~\shortciteA{zambaldiRelationalDeepReinforcement2018,zambaldiDeepReinforcementLearning2018} incorporate a relational inductive bias into the model architecture which aids in generalising along ordinal axes of variation, including extrapolation performance. \shortciteA[SchemaNetworks]{kanskySchemaNetworksZeroshot2017} use an object-oriented and entity-focused architecture, combined with structure-learning methods, to learn logical schema which can be used for backwards-chaining-based planning. These schemas generalise zero-shot to novel state spaces as long as the dynamics are consistent. \shortciteA[VAI]{wangUnsupervisedVisualAttention2021} use unsupervised visual attention and keypoint detection methods to enforce a visual encoder to only encode information relevant to the foreground of the visual image, encoding the inductive bias that the foreground is the only part of the visual input that is important.

\shortciteA{tangNeuroevolutionSelfInterpretableAgents2020} introduce AttentionAgent, which uses neuroevolution to optimise an architecture with a hard attention bottleneck, resulting in a network that only receives a fraction of the visual input. The key inductive bias here is that selective attention is beneficial for optimisation and ZSG\@. Their method generalises zero-shot to unseen backgrounds in CarRacing~\shortcite{brockmanOpenAIGym2016} and VizDoom~\shortcite{Kempka2016ViZDoom,openaiOpenAIGymDoomTakeCoverv02016}. \shortciteA[SensoryNeuron]{tangSensoryNeuronTransformer2021} build on AttentionAgent, adding an inductive bias of permutation invariance in the input space. They argue this is useful for improving ZSG, for a similar reason as before: the attention mechanism used for the permutation-invariant architecture encourages the agent to ignore parts of the input space that are irrelevant. \shortciteA[CRAR]{francois-lavetCombinedReinforcementLearning2018} use a modular architecture combining dynamics learning and value estimation both in a low-dimensional latent space, and show improved performance on a simple maze task.

\shortciteA[SHIFTT]{hillHumanInstructionFollowingDeep2020} and \shortciteA[TransferLanfLfP]{lynchLanguageConditionedImitation2021} both use large pretrained models \shortcite{devlinBERTPretrainingDeep2019,yangMultilingualUniversalSentence2019} to encode natural language instructions for instruction following tasks, tackling reward function variation. They both show improved performance to novel instructions, leveraging the generalisation power of the large pretrained model. This can be seen as utilising domain knowledge to improve zero-shot generalisation to novel goal specifications by incorporating the inductive bias that all the goal specifications will be natural language.

While methods in this area appear dissimilar, they all share the motivation of incorporating specific inductive biases into the RL algorithm. There are several ways of incorporating domain knowledge as an inductive bias. The architecture of the model can be changed to process the variation correctly. If the variation is one to which the policy should be invariant, it can either be removed entirely, or adversarial regularisation can be used to ensure the policy's representations are invariant. More broadly, regularisation or auxiliary losses that encourage the policy to handle this variation correctly can be used. A recommendation to make this body of work more systematic is to use the additional types of structural assumptions discussed in \cref{subsection:addassump} as a starting point for developing algorithms that leverage those assumptions -- many of the works discussed here rely on assumptions that can be classified in those introduced in \cref{subsection:addassump}.
For a deeper discussion of inductive biases see \shortciteA{battagliaRelationalInductiveBiases2018}, and for the original ideas surrounding inductive biases and No Free Lunch theorems see \shortcite{wolpertNoFreeLunch1997}.

\subsubsection{Regularisation and Simplicity}

When we cannot encode a specific inductive bias, standard regularisation can be used. This is generally motivated by a paraphrased Occam's razor: the simplest model will generalise the best. Task-agnostic regularisation encourages simpler models that rely on fewer features or less complex combinations of features. For example, L2 weight decay biases the network towards less complex features, dropout ensures the network cannot rely on specific combinations of features, and the information bottleneck ensures that only the most informative features are used.

\shortciteA{cobbeQuantifyingGeneralizationReinforcement2019} introduced CoinRun, and evaluated standard supervised learning regularisation techniques on the benchmark. They investigate data augmentation (a modified form of Cutout~\shortcite{devriesImprovedRegularizationConvolutional2017}), dropout, batch norm, L2 weight decay, policy entropy, and a combination of all techniques. All the techniques separately improve performance, and the combination improves performance further, although the gains of the combination is minimal over the individual methods, implying that many of these methods address similar causes of worse generalisation. Early stopping can be seen as a form of regularisation, and \shortciteA{adaGeneralizationTransferLearning2021} shows that considering training iteration as a hyperparameter (effectively a form of early stopping) improves performance on some benchmarks. Almost all methods report performance at the end of training, as often early stopping would not be beneficial (as can be seen from the training and testing reward curves), but this is likely an attribute of the specific benchmarks being used, and in the future, we should keep early stopping in mind. 

Several methods utilise information-theoretic regularisation techniques, building on the information bottleneck \shortcite{tishbyDeepLearningInformation2015}. \shortciteA[IBAC-SNI]{iglGeneralizationReinforcementLearning2019} and \shortciteA[IB-annealing]{luDynamicsGeneralizationInformation2020} concurrently introduce methods that rely on the information bottleneck, along with other techniques to improve performance, demonstrating improved performance on OpenAI Procgen, and a variety of random mazes and continuous control tasks, respectively. \shortciteA[RPC]{eysenbachRobustPredictableControl2021} extend the motivation of the information bottleneck to the RL setting specifically, learning a dynamics model and policy which jointly minimises the information taken from the MDP by using information from previous states to predict future states. This results in policies using much less information than previously, which has benefits for robustness and zero-shot generalisation, although this method was not compared on standard ZSG benchmarks to other methods. \shortciteA[SMIRL]{chenReinforcementLearningGeneralization2020} use surprise minimisation to improve the performance of the trained policy, although more rigorous benchmarking is needed to know whether this method has a positive effect.

\shortciteA{songObservationalOverfittingReinforcement2019} show that larger model sizes can lead to \emph{implicit regularisation}: larger models, especially those with residual connections, generalise better, even when trained on the same number of training steps. This is also shown in \shortcite{cobbeQuantifyingGeneralizationReinforcement2019,cobbeLeveragingProceduralGeneration2020}.

\subsubsection{Learning Invariances}
Sometimes we cannot rely on a specific inductive bias or standard regularisation. This is a very challenging setting for RL (and machine learning in general) to tackle, as there is a fundamental limit to performance due to a kind of ``no free lunch'' analogy: we cannot expect a policy to generalise to arbitrary contexts. However, several techniques can help, centred around the idea of using multiple training contexts to \emph{learn} the invariances necessary to generalise to the testing contexts. If the factors of variation within the training contexts are the same as the factors of variation between the training and testing contexts, and the values that those factors take in testing are not far from those in training, then you can use that to learn these factors of variation and how to adapt or be invariant to them.

Much work draws on causal inference to learn invariant features from several training contexts.~\shortciteA[ICP]{zhangInvariantCausalPrediction2020} assume a block MDP structure \shortcite{duProvablyEfficientRL2019} and leverage that assumption to learn a state abstraction that is invariant to irrelevant features, which aids in generalisation performance.~\shortciteA[DBC]{zhangLearningInvariantRepresentations2021} use bisimulation metrics to learn a representation that is invariant to irrelevant visual features, and show that bisimulation metrics are linked to causal inference. \shortciteA[DBC-normed-IR-ID]{kemertasRobustBisimulationMetric2021} improve DBC with a norm on the representation space and the use of intrinsic rewards and regularisation. \shortciteA[PSM]{agarwalContrastiveBehavioralSimilarity2021} suggest limitations of bisimulation metrics and instead propose a policy similarity metric, where states are similar if the optimal policy has similar behaviour in that and future states. They use this metric, combined with a contrastive learning approach, to learn policies invariant to observation variation.

Several approaches use multiple contexts to learn an invariant representation, which is then assumed to generalise well to testing contexts. \shortciteA[IPO]{sonarInvariantPolicyOptimization2020} apply ideas from Invariant Risk Minimization \shortcite{arjovskyInvariantRiskMinimization2020} to policy optimisation, learning a representation which enables jointly optimal action prediction across all domains, and show improved performance over PPO on several visual and dynamics variation environments.~\shortciteA[IAPE]{bertranInstanceBasedGeneralization2020} introduce the \emph{Instance MDP}, an alternative formalism for the ZSG problem, and then motivate theoretically an approach to learn a collection of policies on subsets of the training domains, such that the aggregate policy is invariant to any individual context-specific features which would not generalise. They show improved performance on the CoinRun benchmark~\shortcite{cobbeQuantifyingGeneralizationReinforcement2019} compared to standard regularisation techniques. \shortciteA[LEEP]{ghoshWhyGeneralizationRL2021} also produce an invariant policy across a collection of subsets of the training contexts but motivate this with Bayesian RL\@. Their approach learns separate policies on each subset and then optimistically aggregates them at test time, as opposed to IAPE which uses the aggregate policy during training for data collection and learns the collection of policies off-policy. While these approaches are similar there has been no direct comparison between them.

Other approaches try to learn behaviourally similar representations with less theoretical motivation. \shortciteA[CSSC]{liuCrossStateSelfConstraintFeature2020} use behavioural similarity (similarity of short future action sequences) to find positive and negative pairs for a contrastive learning objective. This auxiliary loss aids in zero-shot generalisation and sample efficiency in several OpenAI Procgen games.~\shortciteA[CTRL]{mazoureCrossTrajectoryRepresentationLearning2021} uses clustering methods and self-supervised learning to define an auxiliary task for representation learning based on behavioural similarity, showing improved performance on OpenAI Procgen. \shortciteA[DARL]{liDomainAdversarialReinforcement2021} uses adversarial learning to enforce the representations of different domains to be indistinguishable, improving performance in visually diverse testing contexts, even when only trained on simple training contexts.~\shortciteA[DRIBO]{fanDRIBORobustDeep2021} uses contrastive learning combined with an information-theoretic objective to learn representations that only contain task-relevant information while being predictive of the future. They show improved performance on both visually diverse domains in DeepMind Control, and in OpenAI Procgen.

\subsubsection{Adapting Online}\label{subsubsection:adaptingonline}

A final way to handle differences between training and testing contexts is to adapt online to the testing contexts. \cref{def:zspt} allows that the policy can adjust online within a single episode, as the policy class includes non-Markovian policies (which can equivalently be viewed as adaptation procedures for producing markovian policies). This adaption has to happen within a single episode and for it to be useful, the adaptation will have to be rapid enough to be useful for improved performance within that episode. Most work on Meta RL, which is traditionally concerned with fast adaptation, assumes access to multiple training episodes in the testing CMDP, violating the zero-shot assumption. However, there are works which can adapt zero-shot. While not being exhaustive, we give a brief overview of this body of work here, focusing on key examples where zero-shot generalisation is evaluated.

Many methods learn a context encoder or inference network, which then conditions either a policy or dynamics model for improved ZSG\@. The inference of this context at test-time can be seen as within-episode online adaptation. \shortciteA[UP-OSI]{yuPreparingUnknownLearning2017} uses Online System Identification to infer the context, which then conditions a policy. \shortciteA[EVF]{yen-chenExperienceEmbeddedVisualForesight2019} learns a context inference network end-to-end with a dynamics model and uses this to adapt to novel objects. \shortciteA[RMA]{kumarRMARapidMotor2021} tackles the sim-to-real problem by training an agent using domain randomisation in simulation, and training a context inference model to condition the policy, similar to \shortciteS{yuPreparingUnknownLearning2017} UP-OSI but with learned context inference. \shortciteA[AugWM]{ballAugmentedWorldModels2021} take a similar idea but work in the offline RL setting, so first train a world model from offline data. They then perform domain randomisation of a specific form in the world model, and then use a hard-coded update rule (enabled by the specific form of domain randomisation) to determine the context to condition the policy, enabling it to adapt zero-shot to downstream tasks. These methods all tackle dynamics variation in continuous control or visual tasks. Often, these methods make use of domain randomisation approaches but aim to have a context-conditioned adaptive policy or model, rather than a policy invariant to all possible contexts.

Some approaches use hard-coded adaptation techniques which do not rely on explicitly inferring a context. \shortciteA[TW-MCL]{seoTrajectorywiseMultipleChoice2020} leverages a multi-headed architecture and multiple-choice learning to learn an ensemble of dynamics models that are selected from during deployment by choosing the one with the highest accuracy. \shortciteA[MOLe]{nagabandiDeepOnlineLearning2019} tackles the online learning problem, keeping a continually updated and expanded collection of dynamics models which are chosen between using non-parametric task inference. In both these works the chosen model plans with model-predictive control, and they show improvements in continuous control tasks. \shortciteA[PAD]{hansenSelfSupervisedPolicyAdaptation2021} uses a self-supervised objective to update the internal representations of the policy during the test episode. They improve performance over standard baselines on visually varying DeepMind Control environments as well as CRLMaze~\shortcite{lomonacoContinualReinforcementLearning2020} (a VizDoom~\shortcite{Kempka2016ViZDoom} 3D navigation environment with visual variation), and on zero-shot generalisation to novel dynamics on both a real and simulated robot arm. \shortcite[GrBAL,ReBAL]{nagabandiLearningAdaptDynamic2019} use meta-RL methods (MAML \shortciteA{finnModelagnosticMetalearningFast2017} and RL$^2$ \shortciteA{duanRlFastReinforcement2016}) to learn to quickly adapt a dynamics model using gradients or recurrence, and then use the adapted model to plan. In this case, it becomes more difficult to draw the line between learned and hard-coded update rules: The gradient update itself is hard-coded, but the initialisation is learned, and in the recurrent case the update is fully learned.

Finally, several methods meta-learn an adaptation function during training. RL$^2$~\shortcite{duanRlFastReinforcement2016,wangLearningReinforcementLearn2017} is a meta RL method where a recurrent network is used, the hidden state of which is not reset at episode boundaries, allowing it to learn and adapt within the recurrent state over multiple episodes. While often compared to methods that require multiple training episodes, this method can often adapt and perform well within a single episode, due to the optimisation approach and architecture. \shortciteA{zintgrafVariBADVeryGood2020} introduce an extension to RL$^2$ called VariBAD based on bayesian RL, where the recurrent network learns to produce latent representations that are predictive of future rewards and previous transitions. This latent representation is used by the policy. \shortciteA[BOReL]{dorfmanOfflineMetaLearning2021} adjusts VariBAD to be usable in an offline setting, improving zero-shot exploration using offline data. \shortciteA[HyperX]{zintgrafExplorationApproximateHyperState2021} improves VariBAD with additional exploration bonuses to improve meta-exploration. \shortciteA{niRecurrentModelFreeRL2021} shows that these simple recurrent methods such as these, with carefully tuned implementations, can compete be improved further, often competing with more specialised algorithms, including in Robust RL and Meta-RL settings. \shortciteA[SNAIL]{mishraSimpleNeuralAttentive2018} modelled fast adaptation as a sequence-to-sequence problem and learns an attention-based architecture which encodes sequences of experience to condition the policy. These methods all show improved performance on ZSG tasks, even though their main focus is on adaptation over longer time horizons.

\subsection{RL-Specific Problems And Improvements}\label{subsection:rlspecific}

The motivations in the previous two sections are mostly equally applicable to supervised learning. However, on top of the problems of generalisation from supervised learning, RL has additional problems which inhibit zero-shot generalisation performance. In this section, we discuss methods targeted at these problems, and also discuss methods that improve ZSG purely through more effective optimisation on the training set in a way that does not overfit (at least empirically).

\subsubsection{RL-specific Problems}\label{subsubsection:rlspecific}

Optimisation in RL has additional issues on top of supervised learning, such as the non-stationarity of the data distribution, and the need to explore. These issues likely interact with generalisation in a non-trivial way. \shortciteA[ITER]{iglTransientNonStationarityGeneralisation2021} shows that the non-stationarity of RL training means that policies learn features that do not generalise well, even if they achieve the same training performance. To address this, they introduce a method to iteratively distil the current policy network into a new policy network with reinitialised weights. This reduces the impact of non-stationarity on the new network, as it is being trained on a more stationary distribution. Other RL-specific optimisation issues likely interact with zero-shot generalisation either positively or negatively, and this area deserves further attention if we are to go beyond techniques copied and adjusted from supervised learning.

\subsubsection{Better Optimisation without Overfitting}

Several works improve ZSG by improving the training performance without overfitting.~\shortciteA{cobbePhasicPolicyGradient2020} introduce Phasic Policy Gradient (PPG), which adjusts the training regime and architecture of PPO such that the policy and value functions use entirely separate networks (rather than just separate heads), which allows the value head to be optimised for longer while not causing the policy to overfit. To recover the benefits of a joint representation, the value network is distilled into an auxiliary value head on the policy network. \shortciteA{raileanuDecouplingValuePolicy2021} build on PPG and introduce Decoupled Advantage Actor Critic (DAAC). They distil an advantage function calculated with GAE into the policy instead of a value function, which further ensures that the policy does not overfit to details that may be predictive of value function but not optimal action selection. They both show improved performance on OpenAI Procgen, demonstrating that value functions can be optimised more strongly than policies. \shortciteA[Sparse DVE]{singhSparseAttentionGuided2021} adjusts the architecture of the value function to allow for a multi-modal output, more closely modelling the true value function given just a visual input. This novel architecture combined with sparsity losses to ensure the value function has the desired properties reduces the variance of value function prediction and improves performance in terms of both return and navigation efficiency in OpenAI Procgen.

Another angle on better optimisation is the use of model-based RL (MBRL). Very little work has applied MBRL to ZSG benchmarks, with \shortciteA{anandProceduralGeneralizationPlanning2021} being an initial example. \shortciteA{anandProceduralGeneralizationPlanning2021} apply MuZero Reanalyse \shortcite{schrittwieserMasteringAtariGo2020,schrittwieserOnlineOfflineReinforcement2021}, a SOTA MBRL method, to OpenAI Procgen \shortcite{cobbeLeveragingProceduralGeneration2020}, showing much-improved performance over SOTA model-free methods at much lower sample complexity. This shows the potential of using MBRL to improve zero-shot generalisation to varying states and observations. The authors also apply MuZero to the meta-learning tasks in Meta-World \shortcite{yuMetaworldBenchmarkEvaluation2019}, although the performance there is not as impressive, showing that generalising to new rewards (as is necessary for Meta-World) is not currently as amenable to MBRL approaches.

While the methods described above do not target ZSG specifically, they improve test-time performance on ZSG benchmarks and so are included here. We hope the field will move towards benchmarks like Procgen being the standard for RL (and not just ZSG), such that in time this work is considered standard RL, rather the ZSG specifically.

\subsection{Discussion}\label{subsection:methoddiscuss}

Having described existing methods for ZSG in RL, and categorised them in \cref{table:methodsone,table:methodstwo}, we now draw some broader conclusions about the field, as well as discuss possible alternative classifications of methods. 

\begin{table}[ht!]
\tiny
\centering
\begin{tabularx}{\textwidth}{c|XXXXX}
\toprule
\textbf{Approach} & \multicolumn{5}{c}{\textbf{Evaluation Variation}}\\
& \textbf{Observation} & \textbf{State} & \textbf{Dynamics} & \textbf{Reward} & \textbf{All} \\ \midrule
Data Augmentation & \textcolor{CBBlue}{\emph{SODA (\shortciteauthor{hansenGeneralizationReinforcementLearning2021})}}, \textcolor{CBOrange}{\texttt{RCAN (\shortciteauthor{jamesSimtoRealSimtoSimDataefficient2019})}}, \textcolor{CBBlue}{\emph{RandFM (\shortciteauthor{leeNetworkRandomizationSimple2020}), InDA,ExDA (\shortciteauthor{koTimeMattersUsing2021}), DrQ (\shortciteauthor{kostrikovImageAugmentationAll2021}), SECANT (\shortciteauthor{fanSECANTSelfExpertCloning2021}), UCB-DrAC (\shortciteauthor{raileanuAutomaticDataAugmentation2021}), PAADA (\shortciteauthor{zhangGeneralizationReinforcementLearning2021}), MixStyle (\shortciteauthor{zhouDomainGeneralizationMixStyle2021}), SVAE (\shortciteauthor{hansenStabilizingDeepQLearning2021})}} & \textcolor{CBBlue}{\emph{UCB-DrAC (\shortciteauthor{raileanuAutomaticDataAugmentation2021}), PAADA (\shortciteauthor{zhangGeneralizationReinforcementLearning2021})}} & & & \\\rowcolor{Gray}
Domain Randomisation & & \textcolor{CBGreen}{MD-SAC (\shortciteauthor{renImprovingGeneralizationReinforcement2020})} & \textcolor{CBGreen}{P2PDRL (\shortciteauthor{zhaoRobustDomainRandomised2020}), DR (\shortciteauthor{tobinDomainRandomizationTransferring2017,pengSimtoRealTransferRobotic2018}), CAD$^2$RL (\shortciteauthor{sadeghiCAD2RLRealSingleImage2017}), ADR (\shortciteauthor{openaiSolvingRubikCube2019}), DDL (\shortciteauthor{wellmerDropoutDreamLand2021})} & & \textcolor{CBGreen}{PCG (\shortciteauthor{risiIncreasingGeneralityMachine2020})} \\
Environment Generation & & \textcolor{CBGreen}{POET (\shortciteauthor{wangPairedOpenEndedTrailblazer2019}), PAIRED (\shortciteauthor{dennisEmergentComplexityZeroshot2021}), E-POET (\shortciteauthor{wangEnhancedPOETOpenEnded2020})} & & & \\\rowcolor{Gray}
Optimisation Objectives & & \textcolor{CBBlue}{\emph{PLR (\shortciteauthor{jiangPrioritizedLevelReplay2021})}} & \textcolor{CBBlue}{\emph{WR$^2$L (\shortciteauthor{abdullahWassersteinRobustReinforcement2019}), RARL (\shortciteauthor{pintoRobustAdversarialReinforcement2017}), (S)(R)(E)-MPO(\shortciteauthor{mankowitzRobustReinforcementLearning2020})}} & & \\
Inductive Biases & \textcolor{CBOrange}{\texttt{AttentionAgent (\shortciteauthor{tangNeuroevolutionSelfInterpretableAgents2020}), VAI (\shortciteauthor{wangUnsupervisedVisualAttention2021}), SensoryNeuron (\shortciteauthor{tangSensoryNeuronTransformer2021})}}, \textcolor{CBBlue}{\emph{DARLA (\shortciteauthor{higginsDARLAImprovingZeroShot2018})}} & \textcolor{CBOrange}{\texttt{NAP (\shortciteauthor{vlastelicaNeuroalgorithmicPoliciesEnable2021}), RelationalRL (\shortciteauthor{zambaldiRelationalDeepReinforcement2018,zambaldiDeepReinforcementLearning2018}), SchemaNetworks (\shortciteauthor{kanskySchemaNetworksZeroshot2017})}}, \textcolor{CBBlue}{\emph{IDAAC (\shortciteauthor{raileanuDecouplingValuePolicy2021}), CRAR (\shortciteauthor{francois-lavetCombinedReinforcementLearning2018})}} & \textcolor{CBOrange}{\texttt{RelationalRL (\shortciteauthor{zambaldiRelationalDeepReinforcement2018,zambaldiDeepReinforcementLearning2018})}} & \textcolor{CBOrange}{\texttt{TransferLangLfP (\shortciteauthor{lynchLanguageConditionedImitation2021}), SHIFTT (\shortciteauthor{hillHumanInstructionFollowingDeep2020})}} & \\
\bottomrule
\end{tabularx}
\caption{\small A table categorising all methods for tackling the ZSG problem in RL, part 1 of 2. The columns represent the type of variation (see \cref{table:environments}) the method is evaluated on, and rows represent the classification in \cref{fig:methods}. Colours (and text styles) represent the main adjustment made by the method: \textcolor{CBGreen}{Green normal-text} methods mainly work by adjusting the training environment, \textcolor{CBOrange}{\texttt{Red monospace-text}} methods mainly work through adjusting the architecture, and \emph{\textcolor{CBBlue}{Blue italic}} methods mainly work through adjusting the objective or loss function (including adding auxiliary losses). While changing the loss often requires an architectural adjustment, and often architectural changes require adjusted losses, we focus on the original motivation of the method.}\label{table:methodsone}
\end{table}

\begin{table}[ht]
\tiny
\centering
\begin{tabularx}{\textwidth}{c|XXXXX}
\toprule
\textbf{Approach} & \multicolumn{5}{c}{\textbf{Evaluation Variation}}\\
& \textbf{Observation} & \textbf{State} & \textbf{Dynamics} & \textbf{Reward} & \textbf{All} \\ \midrule
Regularisation & \textcolor{CBOrange}{\texttt{Implicit Regularisation (\shortciteauthor{songObservationalOverfittingReinforcement2019})}} & \textcolor{CBBlue}{\emph{SMIRL (\shortciteauthor{chenReinforcementLearningGeneralization2020}), Mixreg (\shortciteauthor{wangImprovingGeneralizationReinforcement2020}), IBAC-SNI (\shortciteauthor{iglGeneralizationReinforcementLearning2019})}} & \textcolor{CBBlue}{\emph{RPC (\shortciteauthor{eysenbachRobustPredictableControl2021}), IB-annealing (\shortciteauthor{luDynamicsGeneralizationInformation2020})}} & & \textcolor{CBBlue}{\emph{L2,Dropout,etc. (\shortciteauthor{cobbeQuantifyingGeneralizationReinforcement2019}), EarlyStopping (\shortciteauthor{adaGeneralizationTransferLearning2021})}} \\ \\\rowcolor{Gray}
Learning Invariances & \textcolor{CBBlue}{\emph{DRIBO (\shortciteauthor{fanDRIBORobustDeep2021}), DBC (\shortciteauthor{zhangLearningInvariantRepresentations2021}), DBC-normed-IR-ID (\shortciteauthor{kemertasRobustBisimulationMetric2021})}}, \textcolor{CBBlue}{\emph{DARL (\shortciteauthor{liDomainAdversarialReinforcement2021})}}, \textcolor{CBBlue}{\emph{PSM (\shortciteauthor{agarwalContrastiveBehavioralSimilarity2021}), MISA (\shortciteauthor{zhangInvariantCausalPrediction2020})}} & \textcolor{CBBlue}{\emph{IAPE (\shortciteauthor{bertranInstanceBasedGeneralization2020}), DRIBO (\shortciteauthor{fanDRIBORobustDeep2021}), CTRL (\shortciteauthor{mazoureCrossTrajectoryRepresentationLearning2021}), CSSC (\shortciteauthor{liuCrossStateSelfConstraintFeature2020}), PSM (\shortciteauthor{agarwalContrastiveBehavioralSimilarity2021}), LEEP (\shortciteauthor{ghoshWhyGeneralizationRL2021}), IPO (\shortciteauthor{sonarInvariantPolicyOptimization2020})}} & \textcolor{CBBlue}{\emph{IPO (\shortciteauthor{sonarInvariantPolicyOptimization2020})}} & & \\
Fast Adaptation & \textcolor{CBOrange}{\texttt{PAD (\shortciteauthor{hansenSelfSupervisedPolicyAdaptation2021})}} & \textcolor{CBBlue}{\emph{EVF (\shortciteauthor{yen-chenExperienceEmbeddedVisualForesight2019})}}, \textcolor{CBOrange}{\texttt{SNAIL (\shortciteauthor{mishraSimpleNeuralAttentive2018})}}, \textcolor{CBBlue}{\emph{HyperX (\shortciteauthor{zintgrafExplorationApproximateHyperState2021}), VariBAD (\shortciteauthor{zintgrafVariBADVeryGood2020}), RMA (\shortciteauthor{kumarRMARapidMotor2021}), RL$^2$ (\shortciteauthor{duanRlFastReinforcement2016,wangLearningReinforcementLearn2017})}} & \textcolor{CBBlue}{\emph{EVF (\shortciteauthor{yen-chenExperienceEmbeddedVisualForesight2019})}}, \textcolor{CBOrange}{\texttt{SNAIL (\shortciteauthor{mishraSimpleNeuralAttentive2018})}}, \textcolor{CBBlue}{\emph{HyperX (\shortciteauthor{zintgrafExplorationApproximateHyperState2021}), VariBAD (\shortciteauthor{zintgrafVariBADVeryGood2020})}}, \textcolor{CBOrange}{\texttt{TW-MCL (\shortciteauthor{seoTrajectorywiseMultipleChoice2020})}}, \textcolor{CBBlue}{\emph{UP-OSI (\shortciteauthor{yuPreparingUnknownLearning2017}), BOReL (\shortciteauthor{dorfmanOfflineMetaLearning2021}), GrBAL,ReBAL (\shortciteauthor{nagabandiLearningAdaptDynamic2019}), RMA (\shortciteauthor{kumarRMARapidMotor2021}), RL$^2$ (\shortciteauthor{duanRlFastReinforcement2016,wangLearningReinforcementLearn2017})}}, \textcolor{CBOrange}{\texttt{MOLe (\shortciteauthor{nagabandiDeepOnlineLearning2019}), AugWM (\shortciteauthor{ballAugmentedWorldModels2021}), PAD (\shortciteauthor{hansenSelfSupervisedPolicyAdaptation2021})}} & \textcolor{CBOrange}{\texttt{SNAIL (\shortciteauthor{mishraSimpleNeuralAttentive2018})}}, \textcolor{CBBlue}{\emph{HyperX (\shortciteauthor{zintgrafExplorationApproximateHyperState2021}), BOReL (\shortciteauthor{dorfmanOfflineMetaLearning2021}), VariBAD (\shortciteauthor{zintgrafVariBADVeryGood2020}), RL$^2$ (\shortciteauthor{duanRlFastReinforcement2016,wangLearningReinforcementLearn2017})}} & \\ \rowcolor{Gray}
RL-specific Problems & \textcolor{CBOrange}{\texttt{ITER (\shortciteauthor{iglTransientNonStationarityGeneralisation2021})}} & \textcolor{CBOrange}{\texttt{ITER (\shortciteauthor{iglTransientNonStationarityGeneralisation2021})}} & & & \\
Better Optimisation & \textcolor{CBOrange}{\texttt{Sparse DVE (\shortciteauthor{singhSparseAttentionGuided2021}), PPG (\shortciteauthor{cobbePhasicPolicyGradient2020}), DAAC (\shortciteauthor{raileanuDecouplingValuePolicy2021}), MuZero++ (\shortciteauthor{anandProceduralGeneralizationPlanning2021})}} & \textcolor{CBOrange}{\texttt{Sparse DVE (\shortciteauthor{singhSparseAttentionGuided2021}), PPG (\shortciteauthor{cobbePhasicPolicyGradient2020}), DAAC (\shortciteauthor{raileanuDecouplingValuePolicy2021}), MuZero++ (\shortciteauthor{anandProceduralGeneralizationPlanning2021})}} & & \textcolor{CBOrange}{\texttt{MuZero++ (\shortciteauthor{anandProceduralGeneralizationPlanning2021})}} & \\
\bottomrule
\end{tabularx}
\caption{\small A table categorising all methods for tackling the ZSG problem in RL, part 2 of 2. The columns represent the type of variation (see \cref{table:environments}) the method is evaluated on, and rows represent the classification in \cref{fig:methods}. Colours (and text styles) represent the main adjustment made by the method: \textcolor{CBGreen}{Green normal-text} methods mainly work by adjusting the training environment, \textcolor{CBOrange}{\texttt{Red monospace-text}} methods mainly work through adjusting the architecture, and \emph{\textcolor{CBBlue}{Blue italic}} methods mainly work through adjusting the objective or loss function (including adding auxiliary losses). While changing the loss often requires an architectural adjustment, and often architectural changes require adjusted losses, we focus on the original motivation of the method.}\label{table:methodstwo}
\end{table}

\paragraph{Alternative Classifications.}

We have presented one possible classification of RL methods, but there are of course others. One alternative is to classify methods based on whether they change the architecture, environment or objective of the standard RL approach. This is useful from a low-level implementation perspective of what the differences are between approaches. This approach is not as useful for future researchers or practitioners who hope to choose a ZSG method for a concrete problem they are facing, as there is not a clear mapping between implementation details and whether a method will be effective for a specific problem. We do apply this classification through the colours in \cref{table:methodsone,table:methodstwo}, to emphasise the current focus on adjusting the loss or objective function in current methods. Another approach would be to classify methods based on which benchmarks they attempt to solve, or what specific problem motivated their design. This goes too far in the other direction, grounding methods in exactly the benchmarks they tackle. While practitioners or researchers could try and see which benchmark is most similar to their problem, they might not understand which differences between benchmarks are most important, and hence choose a method that is not likely to succeed. This classification is also less useful in pointing out areas where there is less research being done. Our approach strikes a balance between these two approaches, describing the problem motivations and solution approaches at a high level which is useful for both practitioners and researchers in choosing methods and investigating future research directions.

\paragraph{Strong Generalisation Requires Inductive Biases.}\label{paragraph:stronginductive}

As described in \cref{paragraph:genexpect}, there are hard ZSG problems involving combinatorial interpolation or extrapolation. We want to tackle these problems, as they will occur in real-world scenarios when we have limited contexts to train on, or we know the type of variation but cannot create context-MDPs in the deployment context-MDP set (e.g.~due to limited simulators). To tackle these problems, we need stronger inductive biases targeted towards specific types of extrapolation, as there is unlikely to be a general-purpose algorithm that can handle all types of extrapolation \shortcite{wolpertNoFreeLunch1997}. When doing research tackling extrapolative generalisation, researchers should be clear that they are introducing an inductive bias to help extrapolate in a specific way and be rigorous in analysing how this inductive bias helps. This involves also discussing in which situations the bias may hinder performance, for example in a different setting where extrapolation requires something else.

\paragraph{Going Beyond Supervised Learning as Inspiration.}

Methods for improving generalisation from supervised learning have been a source of inspiration for many methods, particularly for visual variation. This is exactly the variation that happens in computer vision, and hence many methods from that field are applicable. However, non-visual forms of generalisation (i.e.~dynamics, state and reward), while equally important are less studied. These challenges will be specific to RL and interact with other problems unique to RL such as the exploration-exploitation trade-off and the non-stationarity of the underlying data distribution. We hope to see more work in the area of non-visual ZSG, particularly when other hard RL problems are present.

\section{Discussion And Future Work}\label{section:discussion}

In this section we highlight further points for discussion, building on those in \cref{subsection:benchmarkdiscuss} and \cref{subsection:methoddiscuss}, including directions for future research on new methods, benchmarks, evaluation protocols and understanding.

\subsection{Generalisation Beyond Zero-Shot Policy Transfer}\label{subsection:beyondsingle}

This survey focuses on zero-shot policy transfer. We believe this problem setting is a reasonable one that captures many challenges relevant to deploying RL systems. However, there are many important scenarios where zero-shot generalisation is impossible, or the assumptions can be relaxed. We will want to move beyond zero-shot policy transfer if we are to use RL effectively in a wider variety of scenarios.

The most sensible way of relaxing the zero-shot assumption is to move into a continual RL \shortcite[CRL]{khetarpalContinualReinforcementLearning2020} setting: future RL systems will likely be deployed in scenarios where the environment is constantly changing, such that the system needs to adapt continually to these changes. Making progress on this setting will require benchmarks, and we agree with \shortciteA{khetarpalContinualReinforcementLearning2020} that there are not enough good benchmarks for CRL\@. Most benchmarks do not enable testing all the different attributes desired of CRL methods, although \shortciteS{powersCORABenchmarksBaselines2021a} CORA is a good first step. New benchmarks for CRL would also be excellent as benchmarks for ZSG, especially if these benchmarks introduce new environments. We expect that methods built for domain generalisation in RL might be suitable in CRL, as the continual learning setting can be conceptualised as one in which the domain that tasks are within changes over time. This is in contrast to benchmarks that evaluate generalisation to levels sampled from the same distribution as during training, such as OpenAI Procgen \shortcite{cobbeLeveragingProceduralGeneration2020}. Hence, we recommend work on building new CRL benchmarks, to make progress on CRL and ZSG together.

Coupled with the idea of CRL as a more realistic setting for generalisation in RL, we can take inspiration from how humans generalise and what they transfer when generalising, to go beyond transferring a single policy. While humans may not always be able to achieve good results zero-shot on a new task, if the task is related to previously seen tasks, they can reuse previous knowledge or skills to learn the new task faster. This broader notion of generalisation of objects other than a complete policy (e.g.~skills or environment knowledge) will become more relevant when we start to build more powerful RL systems. Hierarchical and multi-task RL are related fields, and methods in these settings often learn subcomponents, modules or skills on source tasks (possibly in an unsupervised manner) which they can then use to increase learning speed and performance when transferred to novel tasks \shortcite{eysenbachDiversityAllYou2018,yangMultiTaskReinforcementLearning2020}. It is likely the capability to transfer components other than a single policy will be useful for future systems, and it would hence be beneficial to have benchmarks that enable us to test these kinds of capabilities. However, this is a challenging request to meet, as defining what these components are, and deciding on a performance metric for these subcomponent transfer benchmarks, are both difficult conceptual problems. We hope to see work in this area in the future.

A final assumption which is almost untouched in RL is that of a fixed action space between training and testing. Recently, \shortciteA{jainGeneralizationNewActions2020} introduced a novel problem setting and framework centred around how to generalise to new actions in RL\@. They introduce several benchmarks for testing methods which generalise to new actions, and a method based on learning an action representation combined with an action-ranking network which acts as the policy at test time. There is very little work in this area, and we do not cover it in this survey, but it presents an interesting future direction for ZSG research.

\subsection{Real World Reinforcement Learning Generalisation}\label{subsection:rwrlg}

\shortciteA{dulac-arnoldEmpiricalInvestigationChallenges2021} propose 9 properties that are characteristic of real-world RL.\footnote{The 9 properties are: limited samples; sensor, action and reward delays; high-dimensional state and action spaces; reasoning about constraints; partial observability; multi-objective or poorly specified rewards; low action latency; offline training; and explainable policies} In thinking about the current set of ZSG benchmarks, these properties are relevant in two ways. First, when applying our methods to the real world, we will have to tackle these problems. Hence, it would be beneficial if ZSG benchmarks had these properties, such that we can ensure that our ZSG methods work in real-world environments. 

Second, several properties are particularly relevant for zero-shot generalisation and the design of new benchmarks: (1) \emph{high cost of new samples}, (2) \emph{training from offline data} and (3) \emph{underspecified or multi-objective reward functions}. We explain each of these and their relation to ZSG below.

\paragraph{Context Efficiency.}\label{paragraph:contextefficiency} Addressing (1), it is likely that the high cost of new samples will also mean a high cost of new environment instances or contexts. This means we want methods that are \emph{context efficient} as well as sample efficient, and hence we require benchmarks and evaluation protocols in which only a few contexts are allowed during training, rather than several 1000s. It is also worth investigating if there is an optimal trade-off (for different costs per sample and per context) between new training samples and new contexts. This line of work would revolve around different possible evaluation metrics based on how many contexts are needed to reach certain levels of generalisation performance. Further, there may be ways of actively selecting new contexts to maximise the generalisation performance while minimising the number of new contexts used, effectively a form of active learning. Evaluating context efficiency will be more computationally expensive, as models will need to be repeatedly trained on different numbers of training contexts, so work which figures out how to evaluate this property more efficiently is also beneficial.

\paragraph{Sim-to-Real and Offline RL.}\label{paragraph:simtorealofflinerl} To tackle (2), two options emerge. The first is relying on good simulations, and then tackling the \emph{sim-to-real} problem, and the second is tackling the offline RL problem directly \shortcite{levineOfflineReinforcementLearning2020}. These approaches might be more or less relevant or applicable depending on the scenario: for example, in many robotics applications, a simulator is available, whereas in healthcare settings it is likely learning from offline data is the only approach possible. Sim-to-real is a problem of domain generalisation. If this direction is most relevant, it implies we should focus on building environments that test for good domain generalisation. Existing work on sim-to-real does address this to some extent, but it would be beneficial to have a fully simulated benchmark for testing sim-to-real methods, as that enables faster research progress than requiring real robots. This is a difficult task and is prone to the possibility of overfitting to the simulation of the sim-to-real problem, but it would be useful as an initial environment for testing sim-to-real transfer.

Offline RL is also a problem of generalisation: a key issue here is generalising to state-action pairs unseen in the training data, and most current methods tackle this by conservatively avoiding such pairs \shortcite{kumarConservativeQLearningOffline2020}. If we had methods to reliably extrapolate to such pairs we could improve offline RL performance.

As well as generalisation improving offline RL, it is likely that future RL deployment scenarios will need to tackle the combination of offline RL and ZSG: training policies offline that then generalise to new contexts unseen in the offline dataset. Current offline RL benchmarks \shortcite{fuD4RLDatasetsDeep2021,gulcehreRLUnpluggedSuite2021} do not measure generalisation in this way, but we believe they should enable us to tackle this combined problem: for example, training on offline data from 200 levels in OpenAI Procgen, and evaluating on the full distribution of levels. If tackling this is infeasible with current methods (as both offline RL and ZSG are hard problems), then a good compromise is to first work on the offline-online setting, where offline RL is used for pretraining, followed by online fine-tuning. Some work has been done in this area \shortcite{nairAWACAcceleratingOnline2021}, but this does not tackle the ZSG problem specifically. Creating benchmarks for evaluating these approaches, where the emphasis is on reducing the length of the online fine-tuning stage and evaluating generalisation after fine-tuning, would move us towards truly offline RL generalisation while still being tractable with current methods.

\paragraph{Reward Function Variation.}\label{paragraph:rewardfunctionvary} It is likely future RL systems will be goal- or task-conditioned, as it will be more efficient to train a general system to do several related tasks than to train a different system for each task. Here, as well as generalising to new dynamics and observations, the trained policies will need to generalise to unseen goals or tasks. The policy will need to be able to solve novel problem formulations, and hence have a more generic problem-solving ability. Benchmarks which address this capability are hence necessary for progress towards more general RL systems (see \cref{subsection:strongervariation}).

Related to challenges of generalisation surrounding reward functions, for many real-world problems designing good reward functions is very difficult. A promising approach is using inverse RL \shortcite[IRL]{aroraSurveyInverseReinforcement2020,ng2000algorithms} to learn a reward function from human demonstrations \shortcite{chenLearningGeneralizableRobotic2021,schmeckpeperReinforcementLearningVideos2020,sermanetTimeContrastiveNetworksSelfSupervised2018,sestiniDemonstrationefficientInverseReinforcement2020}, rather than hand-crafting reward functions for each task. This is often more time-efficient, as demonstrating a task is easier than specifying a reward function for it. There are two generalisation problems here: ensuring the learned reward function generalises to unseen context-MDPs during policy training, and ensuring the trained policy generalises to unseen context-MDPs at test time. The first is an IRL generalisation problem, and the second is the standard problem of generalisation we have considered here. Solving both will be important for ensuring that this approach to training agents is effective.

Work building benchmarks and methods to solve these problems would be valuable future work. Some of these directions could be addressed by combining new evaluation protocols with pre-existing environments to create new benchmarks, rather than requiring entirely new environments to be designed and created.

\subsection{Multi-Dimensional Evaluation Of Generalisation}\label{subsection:multidimgen}

Generalisation performance is usually reported using a single scalar value of test-time performance. However, this does not give us much information with which to compare and choose between methods. While better performance on a benchmark, all else being equal, probably means that a method is more useful, it is usually not clear to what extent the ordering of methods on a benchmark's leaderboard is representative of the hypothetical ordering of those methods on a real-world problem scenario for which we have to choose a method. To alleviate this, performance should be reported on multiple different testing context sets which evaluate different types of generalisation, and radar plots (such as those demonstrated by \shortciteR{osbandBehaviourSuiteReinforcement2020}) can be used to compare methods. This will be more useful for comparing methods in a more realistic way, as well as for practitioners choosing between methods.

Very few environments have the context set required for this type of evaluation, and even those that do would require additional work to create the specific testing context sets. Hence, we recommend that future benchmarks for ZSG are designed to enable this type of evaluation. This requires building environments with controllable context sets as well as PCG components (\cref{paragraph:procgengen}), as well as careful thought to create a variety of testing context sets, ensuring they match important types of generalisation. A first way of splitting up testing context sets might be by the type of variation between training and testing, as well as whether interpolation or extrapolation is required to generalise to that context set. There are likely many other ways, which may be domain-specific or general across many domains.

\subsection{Tackling Stronger Types Of Variation}\label{subsection:strongervariation}

Many of the methods for ZSG tackle observation function or state space variation. Both of these (or at least the practical implementations of them used in the corresponding benchmarks) tend to produce environment families in which it is much simpler to verify (at least intuitively) The Principle Of Unchanged Optimality, meaning that tackling the ZSG problem is tractable; generalisation problems with this style of variation tend to be easier to solve. These two types of variation will appear in real-world scenarios, but the other types of variation are equally important and often harder to tackle.

Work on dynamics is mostly focused on two specific settings: sim-to-real transfer, and multi-agent environments. In sim-to-real transfer \shortcite{zhaoSimtoRealTransferDeep2020}, there is always dynamics variation between the simulator and reality, and much work has focused on how to train policies in this setting. More generally, work on robotics and continuous control tends to address some forms of dynamics variation either in how the robot itself is controlled (e.g.~due to degrading parts) or in the environment (e.g.\ different terrain). In multi-agent environments, if the other agents are considered part of the environment (for example in a single-agent training setting), then varying the other agents varies the dynamics of the environment \shortcite{openendedlearningteamOpenEndedLearningLeads2021}. These both occur in the real world, but there are inevitably other forms of dynamics variation which are less well-studied. Investigating what these other forms of dynamics variation are and whether studying them would be useful is promising future work.

Tackling reward-function variation will be required to train general-purpose policies that can perform a variety of tasks, and generalise to unseen tasks without further training data (as discussed in \cref{paragraph:rewardfunctionvary}). This variation is more difficult to tackle, and it is often difficult or impossible to verify The Principle Of Unchanged Optimality \shortcite{irpanPrincipleUnchangedOptimality2019}. However, we must do work on research to tackle these problems, as otherwise RL approaches will be limited to less-ambitious problems or less-general applications. Further work building more benchmarks that enable testing for reward function variation, especially beyond simple styles of goal specification such as a target state, would be beneficial. Special attention needs to be paid to The Principle Of Unchanged Optimality \shortcite{irpanPrincipleUnchangedOptimality2019} while building these benchmarks: current work tends to handle this by conditioning the policy on a goal or reward specification. Research on what approaches to goal specification are both tractable for policy optimisation and useful for real-world scenarios would be beneficial, as there is likely a trade-off between these two desirable attributes. \shortciteS{hillHumanInstructionFollowingDeep2020,lynchLanguageConditionedImitation2021} work provide good examples of investigating natural language as goal specification, utilising pretrained models to improve ZSG, and we look forward to seeing more work in this area.

\subsection{Understanding Generalisation In Reinforcement Learning}\label{subsection:understandinggen}

While beyond the scope of this survey, several works try to understand the problems underlying generalisation in RL\@. Works in this area include \shortciteS{songObservationalOverfittingReinforcement2019} results, which describe the notion of observational overfitting as one cause of the generalisation gap in RL;~\shortciteA{bengioInterferenceGeneralizationTemporal2020} analyses the relationship between gradient interference and generalisation in supervised and RL showing that temporal difference methods tend to have lower-interference training, which correlates with worse generalisation; \shortciteA{iglTransientNonStationarityGeneralisation2021} studies transient non-stationarity in RL and shows that it negatively impacts RL generalisation; and \shortciteA{hillEnvironmentalDriversSystematicity2020} investigates what environmental factors affect generalisation in an instruction-following task, finding for example that an egocentric viewpoint improves generalisation, as does a richer observation space.

This research is barely scratching the surface of understanding why generalisation in RL in particular is a challenge, and there is much future work to be done. This will enable us to build better methods, and understand any theoretical limits to the diversity of tasks that an RL agent can solve given a limited number of training contexts. The precise empirical experimentation required for this kind of research is exactly that which is enabled by having tight control over the factors of variation in the environments being used, which reinforces the conclusion made in \cref{paragraph:procgengen} that purely PCG environments are unsuited for a study of generalisation in RL\@.

\subsection{Future Work On Methods For Zero-shot Generalisation}\label{subsection:futuremethods}

In this subsection, we summarise directions for future work on methods for ZSG, informed by \cref{section:methods}.

As described in \cref{subsection:understandinggen}, there are many RL-specific factors that interact with generalisation performance, often likely in a negative way. Examples of these factors include the non-stationarity of the data distribution used for training; bootstrapping and TD learning in general; and the need for exploration. Work to understand these factors and then build methods to tackle them as discussed in \cref{subsubsection:rlspecific} is a fruitful direction for future work.

We often have a context space that is unstructured or contains many unsolvable context-MDPs. Methods that enable more effective sampling from these context spaces can alleviate this. Several methods were covered in \cref{subsubsection:environmentgen} but more work in this area, tackling more challenging and realistic environments with different types of variation, would be beneficial.

While much work has been done on meta RL, most work focuses on few-shot adaptation. However, work in this area could be adapted to tackle zero-shot policy transfer settings, if the environment has long episodes that require or enable online adaptation. Enabling the policy to learn and adapt online, and learning this adaptation, would likely improve performance. These approaches would also be more suited to tackling stronger forms of variation (\cref{subsection:strongervariation}), as online adaptation may be necessary in these scenarios. Initial work in this area is described in \cref{subsubsection:adaptingonline}, but much more research should be done.

There are several under-explored approaches that cut across the categorisation in this work. As shown in \cref{table:methodsone,table:methodstwo}, most methods focus on changing the loss function or algorithmic approach. Architectural changes informed by inductive biases are less well studied, with notable examples coming from the work of \shortciteA{cobbePhasicPolicyGradient2020,singhSparseAttentionGuided2021,tangNeuroevolutionSelfInterpretableAgents2020,vlastelicaNeuroalgorithmicPoliciesEnable2021,zambaldiDeepReinforcementLearning2018}. More work can be done on investigating different architectures, either taking inspiration from supervised learning or creating RL-specific architectures. These architectures could encode inductive biases in ways that are difficult to encode through the use of auxiliary losses or regularisation. A second under-explored area is model-based reinforcement learning (MBRL) for ZSG\@. Most methods surveyed here are model-free, with notable exceptions being the work of \shortciteA{ballAugmentedWorldModels2021,kanskySchemaNetworksZeroshot2017,seoTrajectorywiseMultipleChoice2020,anandProceduralGeneralizationPlanning2021}. Learning a world model and combining it with planning methods can enable stronger forms of generalisation, especially to novel reward functions (if the reward function is available during planning). As long as the model generalises well, it could also enable generalisation to novel state and observation functions. World models which can adapt to changing dynamics will be more challenging, but \shortciteA{seoTrajectorywiseMultipleChoice2020} give an initial example. \shortciteA{anandProceduralGeneralizationPlanning2021} is the first example investigating how well standard MBRL approaches generalise, and we look forward to seeing more work in this area.

\section{Conclusion}\label{section:conclusion}

The study of ZSG in RL is still new but is of vital importance if we want to develop applicable and usable RL solutions to real-world problems. In this survey we have aimed to clarify the terminology and formalism concerning ZSG in RL, bringing together disparate threads of research together in a unified framework. We presented a categorisation of benchmarks for ZSG, splitting the taxonomy into environments and evaluation protocols, and we categorised existing methods for tackling the wide variety of ZSG problems.

Here we summarise the key takeaways of this survey (with pointers to the more in-depth discussion of the takeaway). The first two takeaways are concerned with the problem setting as a whole. The next four are focused on evaluating ZSG through benchmarks and metrics, and future work in these areas. The last two are focused on methods for tackling the ZSG problem.
\begin{itemize}
	\item Zero-shot policy transfer is useful to study, even if in specific settings we may be able to relax the zero-shot assumption, as it provides base algorithms upon which domain-specific solutions can be built (\cref{paragraph:0shottransfer}).
	\item However, more work should be done to look beyond zero-shot policy transfer, particularly at continual reinforcement learning, as a way to get around the restriction of the principle of unchanged optimality (\cref{subsection:beyondsingle}).
	\item Purely black-box PCG environments are not useful for testing specific forms of generalisation and are most useful for ensuring robust improvements in standard RL algorithms. Combining PCG and controllable factors of variation is our recommended way to design new environments, having the best trade-off between high variety and the possibility of scientific experimentation (\cref{paragraph:procgengen}). This also enables a more multidimensional approach to evaluating generalisation performance (\cref{subsection:multidimgen}), and specific experimentation aimed at improving our understanding of ZSG in RL (\cref{subsection:understandinggen}).  \item For real-world scenarios, we have to consider both sample efficiency and context efficiency. Evaluating the performance of methods on different sizes of training context sets is a useful evaluation metric which gives us more information to choose between different methods (\cref{paragraph:contextefficiency}).
	\item Work on generalisation problems associated with offline RL is under-explored and would ensure that offline RL approaches are able to generalise effectively (\cref{paragraph:simtorealofflinerl}).
	\item While observation-function and state-space variation are commonly studied, dynamics variation is only tackled in limited settings and reward-function variation is very under-studied. These stronger forms of variation are still likely to appear in real-world scenarios, and hence should be the focus of future research (\cref{subsection:strongervariation}).
	\item For stronger forms of ZSG, stronger inductive biases are necessary, and research should be up-front about what the inductive bias they are introducing is, how it tackles the specific benchmark they are tackling, and how general they expect that inductive bias to be (\cref{paragraph:stronginductive}).
	\item There is much underexplored future work in developing new methods for improved ZSG, such as model-based RL, new architectures, fast online adaptation, solving RL-specific generalisation problems, and environment generation (\cref{subsection:futuremethods}).
\end{itemize}

We hope that this survey will help clarify and unify work tackling the problem of zero-shot generalisation in RL, spurring further research in this area, and serve as a touch-point and reference for researchers and practitioners both inside and outside the field.

\section*{Acknowledgements}

We thank (in alphabetical order) Flo Dorner, Jack Parker-Holder, Katja Hofmann, Laura Ruis, Maximilian Igl, Mikayel Samvelyan, Minqi Jiang, Nicklas Hansen, Roberta Raileanu, Yingchen Xi and Zhengyao Jiang for discussion and comments on drafts of this work. We also thank (alphabetically) Andr{\'e} Biedenkapp, Chelsea Finn, Eliot Xing, Jessica Hamrick, Pablo Samuel Castro, Sirui Xie, Steve Hansen, Theresa Eimer, Vincent Fran{\c{c}}ois{-}Lavet and Zhou Kaiyang for pointing out missing references or work for an updated version of this survey. Finally, we thank Frans Oliehoek and the other reviewers at JAIR for all their helpful criticism and advice, which resulted in a much-improved paper.

\paragraph*{Author Contributions:}\label{section:contributions}
\textbf{Robert Kirk} led the work, developed the formalism, benchmarks categorisation, methods categorisation, and discussion and future work, wrote the full manuscript of the survey, and wrote successive drafts with comments and feedback from the other authors.
\textbf{Amy Zhang} wrote parts of \cref{subsection:backgroundgen,subsection:addassump,subsection:formdiscussion,subsection:benchmarkdiscuss,app:structure}, as well as providing improvements on the entire work through discussion and editing.
\textbf{Tim Rockt\"{a}schel} and \textbf{Edward Grefenstette} advised Robert Kirk, providing discussion and feedback in developing the ideas behind the survey, and provided feedback and comments on the manuscript.

\appendix
\section{Other Structural Assumptions on MDPs}\label{app:structure}
Other forms of structured MDPs have been defined beyond the contextual MDP~\shortcite{hallakContextualMarkovDecision2015} and leveraged to develop algorithms that exploit those structural assumptions. 
One type that holds promise for generalisation is the factored MDP\@.
A \textbf{factored MDP} assumes a state space described by a set of discrete variables, denoted $S:=\{S_1,S_2,\ldots,S_n\}$ \shortcite{boutilierStochasticDynamicProgramming2000,strehlEfficientStructureLearning2007}. We follow the notation and definitions used by \shortciteA{osbandNearoptimalReinforcementLearning2014}. The transition function $T$ has the following property:
\begin{definition}[Factored transition functions]\label{def:factored_transition}
Given two states $s,s'$ and action $a$ in a factored MDP $M$, the transition function satisfies a conditional independence condition
\begin{equation}
    T(s'|s,a)=\prod_i P(s'_i|s,a),
\end{equation}
where $P(s'_i|s,a)$ is the probability distribution for each factor $S_i$ conditioned on the previous state and action. 
\end{definition}
Parallels to causal graphs can be drawn \shortcite{scholkopfCausalityMachineLearning2019}, where a causal graph is a DAG, each vertex is a variable, and the directed edges represent causal relationships. We can rewrite the conditional independence assumption as
\begin{equation}
    T(s'|s,a)=\prod_i P(s'_i|\text{PA}(s'_i),a),
\end{equation}
where the probability of each factor $s_i$ only depends on its parent factors $\text{PA}(s_i)$ from the previous time step. Ideally, these parents are only a subset of all factors so this representation results in a reduction in size from the original MDP\@. Further, this enforces that there are no synchronous edges between factors in the same time step. Critically, the rewards can also be factored in the following way:
\begin{definition}[Factored reward functions]\label{def:factored_reward}
Given two states $s,s'$ and action $a$ in a factored MDP $M$, the reward function satisfies a conditional independence condition
\begin{equation}
    \mathbb{E}\big[R(s,a)\big]=\sum_i \mathbb{E}\big[R_i(s_i,a)\big],
\end{equation}
where $R_i(s_i,a)$ is the reward function for each factor $S_i$. 
\end{definition}
One can think of this factored MDP framework as an extension of the single context-MDP, where the combination of one or more of these factors can be represented as the context, with the space of all possible combinations of factors being the context set. In the latter case, this formulation explicitly encodes how generalisation can be achieved to new contexts via \emph{systematicity} (\cref{subsection:backgroundgen}, \shortciteR{hupkesCompositionalityDecomposedHow2020}): the policy will be trained on contexts taking some values within the set of possible factors, and then be tested on unseen combinations of seen factors.

A more restricted form of structured MDP is the \textbf{relational MDP}~\shortcite{mausam2003solving}. A relational MDP is described by tuple $\langle C,F,A,D,T,R\rangle$. $C$ is the set of object types, $F$ is the set of fluent schemata that are arguments that modify each object type. $A$ is the set of action schemata that acts on objects, $D$ is the set of domain objects, each associated with a single type from $C$, and finally, $T$ is the transition function and $R$ is the reward function. An additional assumption is that objects that are not acted upon do not change in a transition.
Note that the relational MDP can be expanded into a factored MDP that does not assume the additional structure of the form of object types with invariant relations. While this form of MDP is a Planning Domain Definition Language (PDDL) and therefore lends itself well to planning algorithms, it is overly complex for learning algorithms. 

\textbf{Object-oriented MDPs} \shortcite{diukObjectorientedRepresentationEfficient2008} are a simpler form of relational MDPs that are less constrained, and therefore hold more promise for learning methods. Objects, fluents, and actions are defined in the same way as in relational MDPs, but all transition dynamics are determined by a set of Boolean transition terms which consist of a set of pre-defined relation terms between objects and object attributes. In spite of this simplification, it is still significantly constrained compared to CMDPs and can be difficult to use when describing complex systems. 

A final example of a structured MDP is the \textbf{Block MDP}.
Block MDPs~\shortcite{duProvablyEfficientRL2019} are described by a tuple $\langle \mathcal{S}, \mathcal{A}, \mathcal{X}, p, q, R \rangle$ with a finite, unobservable state space $\mathcal{S}$ and possibly infinite, but observable space $\mathcal{X}$. $p$ denotes the latent transition distribution, $q$ is the (possibly stochastic) emission function and $R$ the reward function. This structured MDP is useful in rich observation environments where the given observation space is large, but a much smaller state space can be found that yields an equivalent MDP\@. This allows for improved exploration and sample complexity bounds that rely on the size of that latent state space rather than the given observation space.

\bibliographystyle{theapa}
\bibliography{survey}

\end{document}